\documentclass[a4paper,fleqn]{cas-sc}
\let\printorcid\relax
\usepackage[numbers]{natbib}
\usepackage{pifont}
\usepackage{threeparttable} 
\usepackage{tabularx}
\usepackage{cas-common}
\def\tsc#1{\csdef{#1}{\textsc{\lowercase{#1}}\xspace}}
\tsc{WGM}
\tsc{QE}
\begin{document}
\let\WriteBookmarks\relax
\def\floatpagepagefraction{1}
\def\textpagefraction{.001}

\shorttitle{}    

\shortauthors{}  

\title [mode = title]{LPTR-AFLNet: Lightweight Integrated Chinese License Plate Rectification and Recognition Network}  

\author[1,2]{Guangzhu Xu}[type=editor,
                        auid=000,bioid=1,
                        style=chinese]

\credit{Conceptualization of this study, Methodology development, Designing the experiments, Critical review and editing of the manuscript, Writing – Original Draft Preparation (Primary and Core Sections)}

\affiliation[1]{organization={Hubei Key Laboratory of Intelligent Vision Based Monitoring for Hydroelectric Engineering, China Three Gorges University},
                addressline={}, 
                city={Yichang},
                postcode={443002}, 
                state={Hubei},
                country={China}}

\author[2]{Pengcheng Zuo}[style=chinese]

\credit{Data curation and processing, Conducting formal analysis, Contributing to the methodology and software development, Visualization of results. Writing – Original Draft Preparation (Significant Sections)}

\author[2]{Zhi Ke}[style=chinese]

\credit{Data curation and processing, Validating the experimental results, Contributing to software implementation}

\affiliation[2]{organization={College of Computer and Information Technology, China Three Gorges University}, 
                city={Yichang},
                postcode={443002}, 
                state={Hubei},
                country={China}}

\author[3,4]{Bangjun Lei}[style=chinese]

\credit{Project administration, Critical review and editing of the manuscript, Securing funding for the research}

\affiliation[3]{organization={Hubei Key Laboratory of Digital Finance Innovation}, 
                city={Wuhan},
                postcode={430205}, 
                state={Hubei}, 
                country={China}}
\affiliation[4]{organization={School of Information Engineering, Hubei University of Economics}, 
                city={Wuhan},
                postcode={430205}, 
                state={Hubei}, 
                country={China}}





\begin{abstract}
Chinese License Plate Recognition (CLPR) faces numerous challenges in unconstrained and complex environments, particularly due to perspective distortions caused by various shooting angles and the correction of single-line and double-line license plates. Considering the limited computational resources of edge devices, developing a low-complexity, end-to-end integrated network for both correction and recognition is essential for achieving real-time and efficient deployment. In this work, we propose a lightweight, unified network named LPTR-AFLNet for correcting and recognizing Chinese license plates, which combines a perspective transformation correction module (PTR) with an optimized license plate recognition network, AFLNet. The network leverages the recognition output as a weak supervisory signal to effectively guide the correction process, ensuring accurate perspective distortion correction. To enhance recognition accuracy, we introduce several improvements to LPRNet, including an improved attention module to reduce confusion among similar characters and the use of Focal Loss to address class imbalance during training. Experimental results demonstrate the exceptional performance of LPTR-AFLNet in rectifying perspective distortion and recognizing double-line license plate images, maintaining high recognition accuracy across various challenging scenarios. Moreover, on lower-mid-range GPUs platform, the method runs in less than 10 milliseconds, indicating its practical efficiency and broad applicability. \nocite{*}
\end{abstract}


\maketitle
\section{Introduction}\label{sec:introduction}
In recent years, with the continued growth in vehicle ownership, Automatic License Plate Recognition (ALPR) systems have found widespread application in traffic management, parking management, security surveillance, intelligent transportation, and law enforcement. In constrained environments characterized by dense vehicle populations, such as parking lots and toll booths, integrated barrier systems and stable light sources are commonly deployed to ensure high-quality acquisition of license plate images, thereby enabling efficient and accurate recognition. However, in unconstrained environments with variable lighting conditions, diverse shooting angles, and complex weather conditions, license plate image quality is often poor, exhibiting issues such as blurring, occlusion, and skewing, which significantly reduce the accuracy of license plate localization and character recognition. These image degradation phenomena increase the difficulty of license plate recognition in unconstrained environments, posing a significant challenge for current research\cite{liu2024improving}\cite{kim2024afa}\cite{dhyani2023real}.

License plate recognition systems typically comprise three key components: license plate detection, rectification, and character recognition. While notable advancements have been achieved in license plate detection and character recognition algorithms, research on license plate rectification remains relatively limited. Presently, mainstream rectification methods can be categorized into two types: the first type is based on the localization of the four vertices of the license plate, employing perspective transformation for image correction \cite{gautam2023deep}\cite{kundrotas2023two}. This approach offers advantages such as high computational efficiency and straightforward implementation; however, its performance heavily depends on the accuracy of vertex localization. Errors in vertex positioning can lead to rectification distortions, subsequently degrading character recognition performance, and this method also demands high-precision annotation of training data. The second category involves estimating spatial transformation parameters for correction, exemplified by the Spatial Transformation Network (STN)\cite{jaderberg2015spatial}, which is often integrated with license plate recognition models and trained in an end-to-end manner to enhance rectification quality\cite{zhang2022research}\cite{shao2023multi}\cite{deng2023spatial}. Nevertheless, STN exhibits limitations when handling license plates with significant perspective distortion, primarily because it is better suited for affine transformations and has limited adaptability to nonlinear deformations. Besides STN, deformable convolutional networks (DCN)\cite{wang2023internimage}have also been employed for license plate rectification\cite{liu2024irregular}\cite{srinivasan2024real}. Although DCN introduces learnable offsets that allow convolution kernels to adaptively adjust their sampling locations—potentially improving deformation handling—experimental results indicate that, for license plates viewed nearly frontally, the offset adjustments introduced by DCN may induce unnecessary deformations, thereby counteracting recognition accuracy.

Current state-of-the-art license plate recognition algorithms predominantly favor segmentation-free character recognition approaches. These methods take the entire license plate image as input, leveraging convolutional neural networks (CNNs) or convolutional recurrent neural networks (CRNNs) for end-to-end feature learning and recognition\cite{seo2022robust}\cite{gong2022unified}\cite{peng2023end}\cite{10458015}. While these methods offer advantages in terms of accuracy, they typically rely on high-performance GPU hardware to achieve real-time processing, thereby limiting their deployment in practical application scenarios. Consequently, low-computational-cost, lightweight license plate recognition models are urgently needed for real-world applications, particularly within intelligent transportation systems and edge computing devices, where such models are better suited to meet requirements for high efficiency and low power consumption.

Compared to recognition methods based on CRNNs\cite{li2206pp}, lightweight license plate recognition models centered on CNNs support highly parallelized computation, substantially accelerating training speed. LPRNet\cite{zherzdev2018lprnet} is one of the few purely end-to-end license plate recognition models that combines CNN and Connectionist Temporal Classification (CTC) techniques. It has a relatively small number of parameters (only about 467K), which grants it good adaptability for edge devices. However, research by Zou et al. \cite{zou2022license}indicates that LPRNet's accuracy on the deformed subset of the CCPD single-line license plate dataset—specifically regarding rotations and tilts—still has room for improvement compared to more complex models. This limitation primarily arises from LPRNet's insufficient utilization of spatial positional information of characters. In practical recognition scenarios, the lack of character spatial context can lead to feature confusion and character adhesion, thereby reducing recognition accuracy.

In unrestricted environments, Chinese license plate recognition faces the challenge of handling both single-line and double-line plates, which are commonly encountered. To achieve license plate rectification informed by recognition results through the linkage of rectification and recognition networks, the recognition model needs to process both single and double-line plates simultaneously. Currently, robust solutions addressing this requirement remain underdeveloped. Furthermore, the scarcity of high-quality double-line Chinese license plate datasets, particularly in unconstrained environments, significantly hinders the performance improvement of models across all stages of license plate recognition. While publicly available datasets like CCPDv1\cite{xu2018towards} and CCPDv2\cite{timmurphy.org2} offer diversity in shooting environments and angles, they primarily feature single-line Anhui province license plates, lacking the necessary variety to effectively train models for unconstrained scenarios.

To address the aforementioned challenges, this paper proposes a lightweight integrated rectification and recognition network for both single-line and double-line Chinese license plates. To overcome the issue that double-line plates cannot be directly recognized—leading to difficulties in providing effective supervision signals for the recognition network—we extend the rectification network by supervising it with double-line plate images recognized by the single-line recognition network after correction. Additionally, in response to the scarcity of double-line license plate datasets, this paper constructs a dedicated double-line license plate dataset. Regarding model performance, to improve upon the limitations of LPRNet, we introduce a lightweight per-channel attention (LP-CA) module and adopt Focal CTC\cite{shah2019sangctc}, aiming to enhance recognition accuracy while maintaining its real-time processing speed.

The main contributions of this paper are as follows:

A lightweight Perspective Transformation Rectification (PTR) module is introduced for automatic rectification of single-line license plate images. This module innovatively combines license plate vertex coordinate estimation with inverse perspective transformation, thereby eliminating the need for direct regression of perspective transformation matrix parameters, a common practice in traditional methods. This design effectively mitigates the inherent instability challenges encountered by conventional Spatial Transformer Networks (STNs) in regressing perspective matrix parameters. Furthermore, by optimizing PTR jointly with a license plate recognition network, self-adaptive license plate rectification is achieved without requiring explicit annotations, significantly enhancing the system's practicality and adaptability.

To overcome the performance limitations of the existing lightweight License Plate Recognition Network (LPRNet), an enhanced LPRNet architecture is proposed. Specifically, a lightweight per-channel attention (LP-CA) module is integrated into LPRNet to accentuate the high-level features of license plate characters, thereby reducing character misidentification. Additionally, the traditional CTC loss is replaced with Focal CTC loss, which effectively addresses the challenge of imbalanced Chinese character distribution within the dataset, leading to a significant improvement in LPRNet's recognition performance.

Extending the proposed single-line PTR module, the synchronized spatial rectification of double-line license plates is achieved. This is accomplished by concatenating their upper and lower character regions to effectively transform them into a single-line format, enabling unified distortion rectification for both single-line and double-line license plates while preserving lightweight characteristics. Building upon these advancements, an end-to-end License Plate Transformation and Recognition - Adaptive Feature Learning Network (LPTR-AFLNet) is further constructed. This novel framework integrates the PTR module and the improved LPRNet for collaborative optimization, thereby achieving unified rectification and recognition of both single-line and double-line Chinese license plates within a single, coherent system.

The remainder of this paper is structured as follows: Section \ref{sec:related-works} reviews existing license plate rectification and recognition algorithms, summarizing their respective advantages and disadvantages. Building upon this, Section \ref{sec:method} details the innovative algorithm proposed herein. Section \ref{sec:Experiment-settings-and-Results-Analysis} then presents the experimental results, demonstrating the effectiveness of this algorithm through comparison with established methods. The paper concludes in Section \ref{sec:Conclusion}, summarizing key findings and discussing potential avenues for future research.
\section{Related works}\label{sec:related-works}
\subsection{License plate image rectification}\label{subsec:license-plate-image-rectification}
Geometric correction of license plate images can effectively mitigate variations in character spatial distribution caused by multi-angle shooting, thereby enhancing recognition performance. References \cite{gautam2023deep}\cite{kundrotas2023two} propose correction methods based on license plate localization results, which involve constructing the coordinates of four corner points and utilizing an inverse perspective transformation matrix to perform image correction. This approach offers advantages such as simplicity of algorithm and high computational efficiency. However, it also has two main limitations: First, the correction quality heavily depends on the accuracy of localization, making it highly sensitive to localization errors, which can be influenced by annotation quality. Second, the single correction module lacks a collaborative optimization mechanism with the character recognition system, making it difficult to adaptively adjust based on the specific requirements of the recognition task.

To address these limitations, Reference\cite{zherzdev2018lprnet} introduces Spatial Transformer Networks (STN)\cite{jaderberg2015spatial} for license plate geometric correction. This STN comprises three components: a localization network, a grid generator, and a sampler. The localization network extracts affine transformation parameters through multi-level convolutional and fully connected layers. The grid generator then generates a coordinate grid based on these parameters. Finally, the sampler uses the grid to sample the input image, obtaining the corrected image. Guided by feedback from the license plate recognition system, the model continually optimizes its parameters through supervised signals. Subsequent works \cite{ding2024spatial}\cite{yang4184430robust} have built upon this foundation, developing Spatial Correction Networks (SPN). However, these methods, when regressing the spatial transformation matrix, only consider affine transformations, such as cropping, rotation, and scaling, which are 2D transformations. This poses a limitation in real-world shooting environments, where license plates can also undergo perspective distortion (as illustrated in Figure\ref{fig17}) in addition to basic planar transformations. Consequently, STNs designed solely for affine transformations struggle to handle complex spatial deformations. Although STNs theoretically possess the capability to perform perspective transformations, directly extending them to regress perspective transformation parameters can easily lead to training non-convergence and potentially unstable correction results due to the entangled relationships between multiple transformation parameters.

Addressing the issues with STNs, Reference\cite{xiao2021robust} proposes a Spline-Plate-based Spatial Transformer Network for correcting non-regular regions. This method estimates up to 110 reference points to calculate spline-plate transformation parameters, thereby achieving perspective correction of license plate images. However, estimating a large number of reference points leads to a decrease in computational efficiency and can also cause fluctuations and instability in character positions. 

Deformable Convolution v3 (DCNv3), as proposed in\cite{wang2023internimage}, can be viewed as a network that indirectly achieves spatial transformation. At its core, DCNv3 introduces an offset prediction network within the convolutional operation. This network dynamically predicts offsets for each location, and these offsets are used to adjust the sampling points, enabling irregular sampling of the feature map. As a result, the convolution kernel becomes more adaptable to the shapes and positional variations of the input features, effectively achieving spatial transformation. While this deformable convolution alleviates spatial adaptation challenges to some extent in license plate recognition, its performance remains inferior to STN (Spatial Transformer Network) when dealing with scenarios involving significant geometric distortions or large tilt angles. Furthermore, in situations where the license plate is viewed from a frontal perspective, unstable fluctuations in the convolution offsets can paradoxically lead to a degradation in recognition performance.  
\subsection{License plate recognition}\label{subsec:license-plate-recognition}
Currently, the main approaches for license plate recognition primarily rely on sequence recognition models based on CRNN or pure CNN architectures. CRNN-based license plate recognition models\cite{gong2022unified}\cite{peng2023end} use RNNs (such as LSTM or GRU) to model text sequences, capturing the contextual information between characters. However, CRNNs face limitations in computational efficiency, as their training is relatively slow. Additionally, due to their inherent sequential dependencies, LSTM and GRU are difficult to parallelize, resulting in higher computational complexity. These factors make it challenging to achieve efficient performance in resource-constrained environments.

Compared to this, lightweight license plate recognition models based on pure CNNs\cite{seo2022robust}\cite{10458015}\cite{zherzdev2018lprnet} offer higher computational efficiency. In \cite{zherzdev2018lprnet}, a lightweight CNN-based license plate recognition model called LPRNet is proposed. This model uses a 13×1 vertical convolution kernel to extract vertical contextual features related to character positions (such as stroke structures). Its core mechanism maps the width dimension of the feature map to character sequence positions, while the height dimension encodes local character features. By incorporating multi-scale feature fusion layers and the CTC loss function, the model achieves end-to-end recognition without requiring character segmentation. However, LPRNet has two main limitations: first, it is less robust to large-angle tilts or perspective distortions of the license plate; second, in cases of character merging or tight spacing, the lack of explicit segmentation mechanisms can lead to feature interference. These shortcomings restrict its application in complex scenarios, and future research should focus on improving its adaptability to geometric deformations and its ability to decouple character features.

In recent years, encoder-decoder-based OCR methods have shown potential in license plate recognition. For example, \cite{wang2022lsv} uses a CNN-based feature encoder to extract features from the license plate image, combined with a dedicated OCR model, DAN \cite{wang2020decoupled}, as a decoupled decoder that integrates features with attention maps to generate predictions. This attention mechanism helps the network focus on different character locations during decoding. However, the decoupled decoder has high computational complexity, which reduces inference speed for longer sequences. Additionally, this approach relies heavily on large-scale, high-quality labeled data for training to achieve high accuracy. PP-OCRV3 \cite{li2206pp}, a lightweight OCR model, performs well for general text detection and recognition but has limitations for license plates. It is designed for complex scene detection and deformed text, without specific optimization for fixed regions, short character sequences (e.g., 7 characters), or real-time requirements. Its inclusion of extra modules increases computational load, and its training data lacks focus on license plate formats, resulting in lower accuracy under challenging conditions such as tilt and lighting changes. Consequently, PP-OCRV3's lightweight advantages are not fully utilized, making it unsuitable for direct deployment in license plate recognition tasks.

Current license plate recognition research mainly focuses on single-line plates, with less attention to double-line plates. \cite{qin2020efficient} adds a classification branch for plate type (single/double) to LPRNet, using a "slicing parameter" to divide double-line plates. This parameter relies on a frontal view; tilt causes segmentation errors and affects recognition. To handle single/double-line differences, the recognition head uses a three-branch design (single-line/double-line upper/double-line lower). However, with fixed input size, double-line characters are smaller, and there are scale differences between single and double-line characters (e.g., larger characters on the bottom of single-line plates), increasing feature extraction difficulty. To address this, \cite{shah2019sangctc} replaces the backbone with a lightweight Small Inception Block, improving character scale adaptability via multi-scale convolution, but sacrificing the computational efficiency of LPRNet's original feature extraction. Inception's parallel multi-scale fusion may also weaken the original network's targeted hierarchical feature extraction, potentially impacting recognition robustness. Despite these trade-offs, this work provides a practical approach for unified single/double-line license plate recognition and enhances model generalization through a multi-branch structure and scale-adaptive design.

Double-line license plate recognition in unconstrained environments lags significantly behind its single-line counterpart, primarily due to the scarcity of large, publicly available datasets. Current research in this area predominantly relies on synthetic images or limited private data. The restricted scale and diversity of these datasets prevent them from adequately capturing the wide range of real-world license plate variations and complex backgrounds. This limitation consequently impairs the robustness and generalization capabilities of proposed models in true uncontrolled settings. Therefore, the lack of diverse public data resources constitutes a major bottleneck for the practical deployment of double-line license plate recognition technology.
\section{Method}\label{sec:method}
While lightweight license plate recognition models, exemplified by LPRNet, have achieved significant progress, challenges remain in effectively handling license plates with perspective distortion, accommodating diverse plate formats, and mitigating feature overlap and confusion during recognition. To address these issues, we propose a novel, lightweight, integrated rectification and recognition network for single and double-line Chinese license plates – LPTR-AFLNet. This architecture aims to further enhance the accuracy and robustness of license plate recognition models.

\begin{figure}
  \centering
    \includegraphics[width=0.9\linewidth]{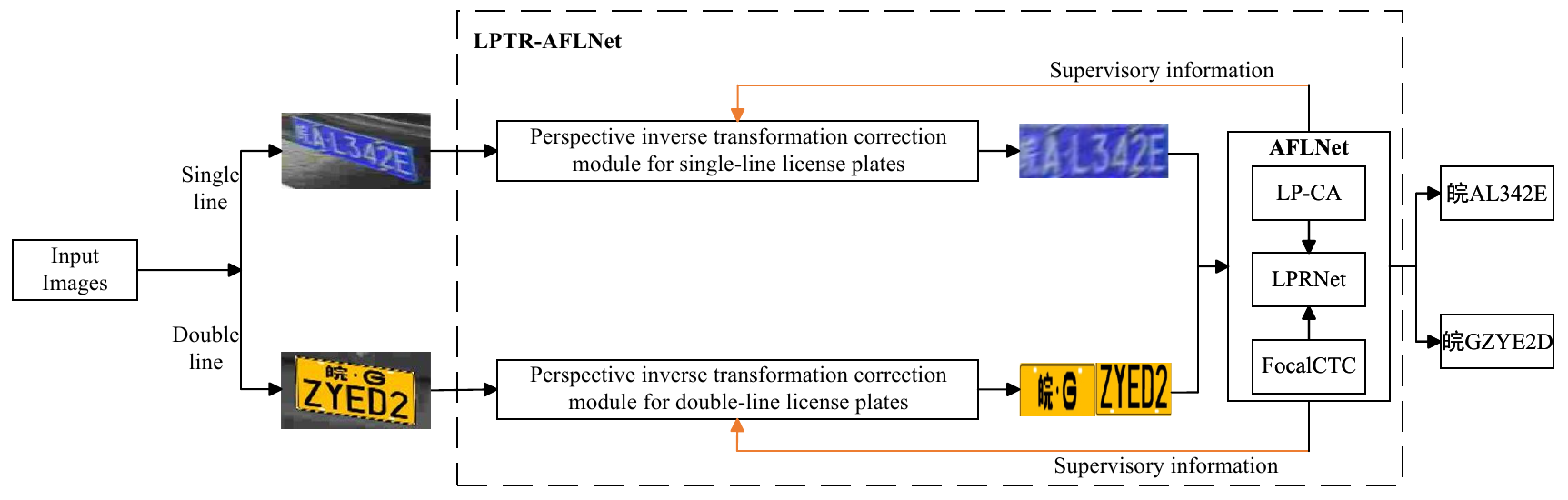}
    \caption{Overall System Block Diagram}\label{fig1}
\end{figure}

As shown in Figure\ref{fig1}, the network supports two types of inputs: one is a license plate image roughly localized via rectangular bounding boxes, and the other is a precisely localized license plate image obtained through vertex-based perspective transformation. For single-line plates, an inverse perspective transformation (IPT) module is used to rectify the license plate into a frontal, recognizable view. For double-line plates, a specially designed perspective transformation module is employed to independently rectify the upper and lower lines, which are then horizontally concatenated to form a unified single-line format. This elegant design ensures that the rectification process for both single- and double-line plates is uniformly handled by the perspective inverse transformation (PTR) module, thereby simplifying the overall workflow. Finally, the rectified license plate images are fed into the lightweight recognition network AFLNet, which is optimized with the proposed channel-wise decoupled attention mechanism and Focal CTC loss to enhance recognition performance.
\subsection{Perspective Transformation Rectification Module for Single-Line License Plates}\label{subsec:Perspective-Transformation-Rectification-Module-fo-Single-Line-License Plates}
\begin{figure}
  \centering
    \includegraphics[width=0.9\linewidth]{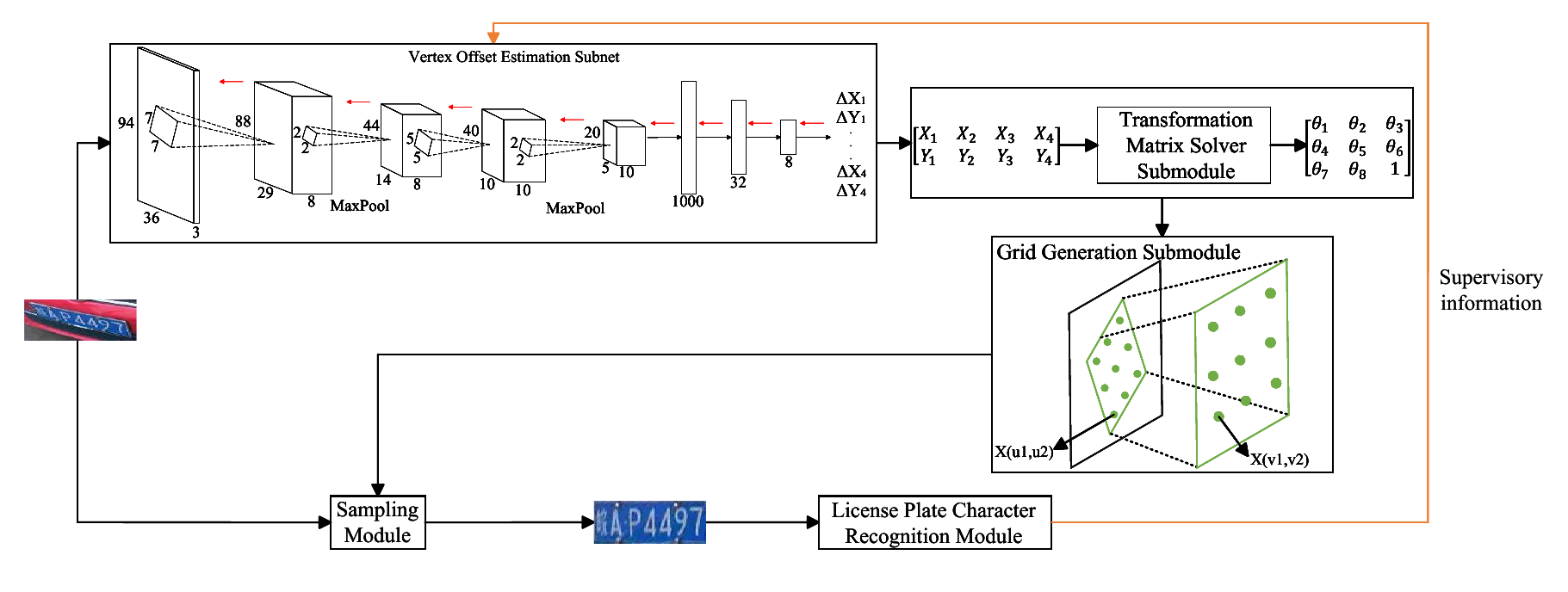}
    \caption{Network structure of perspective distortion rectification module for single-line license plate}\label{fig2}
\end{figure}

The Spatial Transformer Network (STN) is unsuitable for license plate images exhibiting perspective distortion because it only considers affine transformations when regressing the spatial transformation matrix. To address this limitation, this paper proposes a perspective distortion spatial rectification module based on weakly supervised information from license plate recognition. The designed rectification algorithm replaces the localization network in the original STN with a license plate vertex offset estimation sub-network, which facilitates the convergence of the spatial rectification network. Furthermore, a transformation matrix solving sub-module is introduced to directly estimate the parameters of the transformation matrix, effectively resolving the difficulty of accurately regressing perspective transformation parameters when dealing with complex license plate image features in the original STN. The entire network primarily consists of four components: a license plate vertex offset estimation sub-network, a transformation matrix solving sub-module, a grid generation sub-module, and a sampling sub-module. The overall structure is illustrated in Figure\ref{fig2}.
\subsubsection{Vertex Offset Estimation Subnet for License Plates}\label{subsubsec:Vertex-Offset-Estimation-Subnet-for-License-Plates}
\begin{figure}
  \centering
    \includegraphics[width=0.9\linewidth]{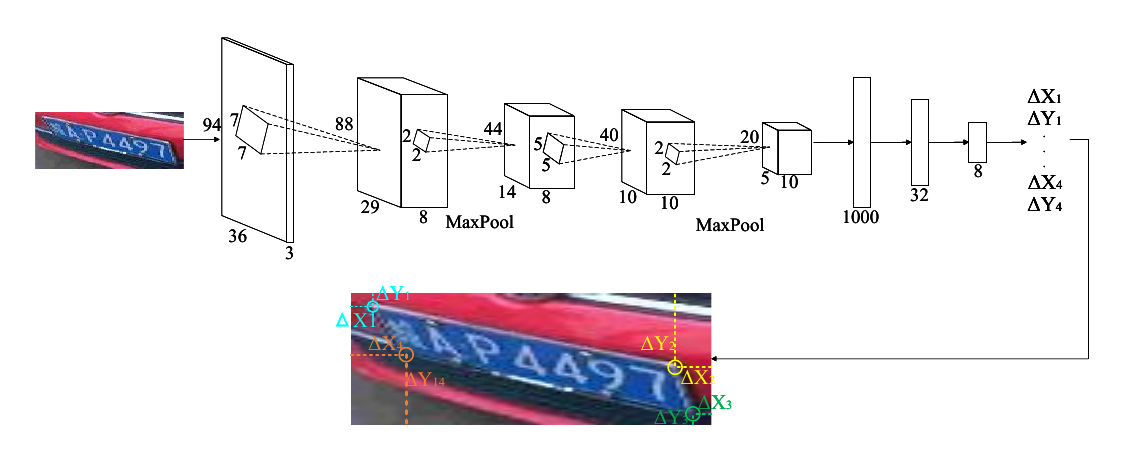}
    \caption{Structure of license plate vertex offset estimation subnetwork}\label{fig3}
\end{figure}

The license plate vertex offset estimation sub-network consists primarily of four convolutional layers and three fully connected layers, as depicted in Figure\ref{fig3}. The convolutional layers are responsible for extracting key features related to the vertex coordinate offsets of the license plate, while the final three fully connected layers are used to regress the required offset values. This entire sub-network regresses eight parameters, representing the offsets of the four license plate vertices with respect to the coordinates of the four corners of the license plate image. As illustrated in Figure\ref{fig3}, $(\Delta X_i, \Delta Y_i)\ (i=1,2,3,4)$ represent the offsets of the license plate vertices relative to the corner coordinates of the license plate image.
\subsubsection{Transformation Matrix Solver Submodule}\label{subsubsec:Transformation-Matrix-Solver-Submodule}
This submodule calculates the four-vertex coordinates of the license plate region. It achieves this by adding the vertex offset values regressed by the vertex offset estimation submodule to the normalized coordinates of the four corner points of the license plate image. Specially,$(X_1,Y_1)=(0,0)+(\Delta X_1,\Delta Y_1),(X_2,Y_2)=(1,0)+(\Delta X_2,\Delta Y_2),(X_3,Y_3)=(1,1)+(\Delta X_3,\Delta Y_3),(X_4,Y_4)=(0,1)+(\Delta X_4,\Delta Y_4)$where $(0,0),(1,0),(1,1)$ and $(0,1)$ represent the normalized coordinates of the four corner points of the license plate image, respectively.

\begin{figure}
  \centering
    \includegraphics[width=0.8\linewidth]{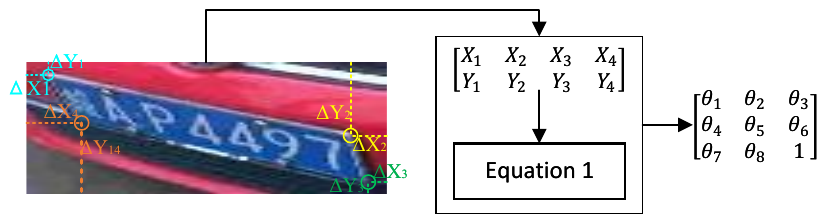}
    \caption{Specific Flowchart of the Transformation Matrix Solver Module}\label{fig4}
\end{figure}

The goal of license plate rectification is to transform the license plate region that has undergone geometric spatial transformation into a rectangular image. In this paper, based on the algorithm\cite{zhang1999flexible}, by using the four sets of corresponding points composed of $(X_i,Y_i)(i=1,2,3,4)$ and the normalized coordinates $(U_i,V_i)(i=1,2,3,4)$ of the four corner points of the rectified output image, Formula 1 is derived. Solving it can obtain the first 8 parameters $(\theta_1 \sim \theta_8)$ of the perspective transformation matrix. The 9th parameter of the perspective transformation matrix is a scale factor, which is set to 1 here. The specific steps of the whole process are shown in Figure \ref{fig4}.
\begin{equation}
\begin{bmatrix}
   U_1 & V_1 & 1 & 0 & 0 & 0 & -X_1U_1 & -X_1V_1 \\ 
   0 & 0 & 0 & U_1 & V_1 & 1 & -Y_1U_1 & -Y_1V_1 \\
   U_2 & V_2 & 1 & 0 & 0 & 0 & -X_2U_2 & -X_2V_2 \\ 
   0 & 0 & 0 & U_2 & V_2 & 1 & -Y_2U_2 & -Y_2V_2 \\
   U_3 & V_3 & 1 & 0 & 0 & 0 & -X_3U_3 & -X_3V_3 \\ 
   0 & 0 & 0 & U_3 & V_3 & 1 & -Y_3U_3 & -Y_3V_3 \\
   U_4 & V_4 & 1 & 0 & 0 & 0 & -X_4U_4 & -X_4V_4 \\ 
   0 & 0 & 0 & U_4 & V_4 & 1 & -Y_4U_4 & -Y_4V_4 \\
\end{bmatrix}
\begin{bmatrix}
   \theta_1 \\  
   \theta_2 \\ 
   \theta_3 \\ 
   \theta_4 \\
   \theta_5 \\
   \theta_6 \\
   \theta_7 \\
   \theta_8 \\
\end{bmatrix}
=
\begin{bmatrix}
   X_1 \\  
   Y_1 \\ 
   X_2 \\  
   Y_2 \\
   X_3 \\  
   Y_3 \\
   X_4 \\  
   Y_4 \\
\end{bmatrix}\label{equation1}
\end{equation}
\subsubsection{Grid Generation Submodule and Sampling Module}\label{subsubsec:Grid Generation-Submodule-and-Sampling-Module}
\begin{figure}
  \centering
    \includegraphics[width=0.8\linewidth]{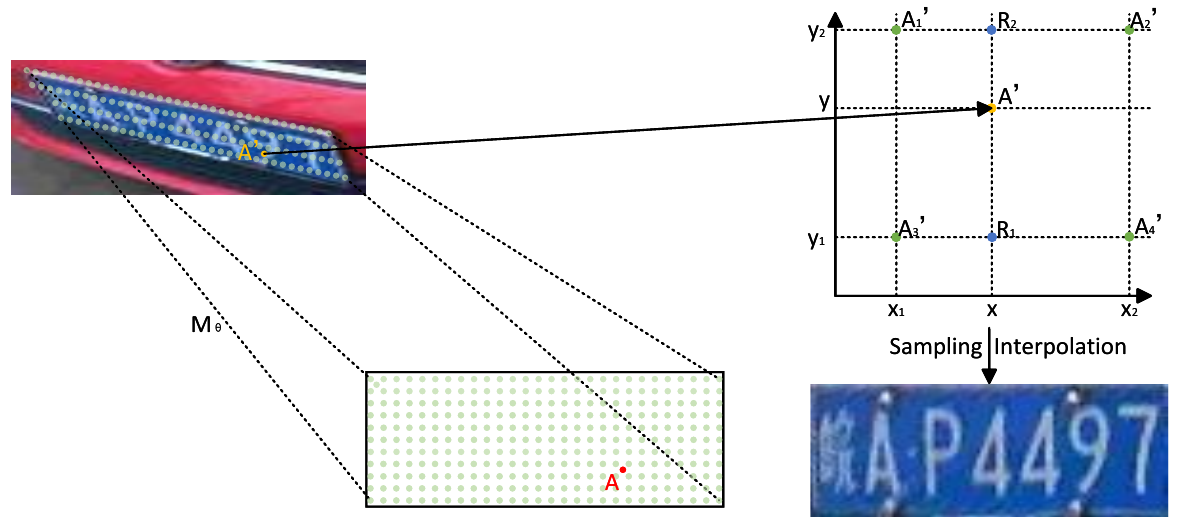}
    \caption{Grid Generation and Sampling Diagram}\label{fig5}
\end{figure}

The grid generation submodule generates a coordinate grid corresponding to the output image, as shown by the green mesh in the Figure\ref{fig5}. Each integer coordinate point in the coordinate grid corresponds to the position of each pixel in the output image. This module constructs a transformation matrix using the transformation parameters calculated by Formula 1, and maps each coordinate point in the output image coordinate grid to the input image to obtain its spatial position in the input image. Since the coordinates of the corresponding positions in the input image may be non-integer values, the sampling module uses bilinear interpolation. Based on the coordinate mapping positions in the input image, it interpolates the pixel values of the corresponding positions in the output image, as shown in the upper right part of Figure\ref{fig5}. Finally, the rectified image of the license plate region in the input image is obtained, as shown in the lower right part of Figure \ref{fig5}. The interpolation calculations are shown in Formulas \ref{equation2}, \ref{equation3} and \ref{equation4}. Here,
$A_1'(x_1,y_2),A_2'(x_2,y_2),A_3'(x_1,y_1),A_4'(x_2,y_1)$are the four integer-coordinate pixel points adjacent to the mapped point $A'$  in the original input image, and their pixel values are $f(A_1'),f(A_2'),f(A_3'),f(A_4')$. First, use Formulas \ref{equation2} and \ref{equation3} to obtain the interpolation results 
$f(x,y_1),f(x,y_2)$ in the horizontal direction. Then, perform linear interpolation in the vertical direction on this basis to obtain the final pixel value $f(x,y)$ of point $A'$.
\begin{equation}
f(x,y_1)\approx\frac{x_2-x}{x_2-x1}f(A_1')+\frac{x-x_1}{x_2-x1}f(A_2')\label{equation2}
\end{equation}
\begin{equation}
f(x,y_2)\approx\frac{x_2-x}{x_2-x1}f(A_3')+\frac{x-x_1}{x_2-x1}f(A_4')\label{equation3}
\end{equation}
\begin{equation}
f(x,y)\approx\frac{y_2-y}{y_2-y_1}f(x,y_1)+\frac{y-y_1}{y_2-y_1}f(x,y_2)\label{equation4}
\end{equation}
\subsection{Perspective Transformation Rectification Module for Double-Line License Plates}\label{subsec:Perspective-Transformation-Rectification-Module-for-Double-Line-License-Plates}
\begin{figure}
  \centering
    \includegraphics[width=0.9\linewidth]{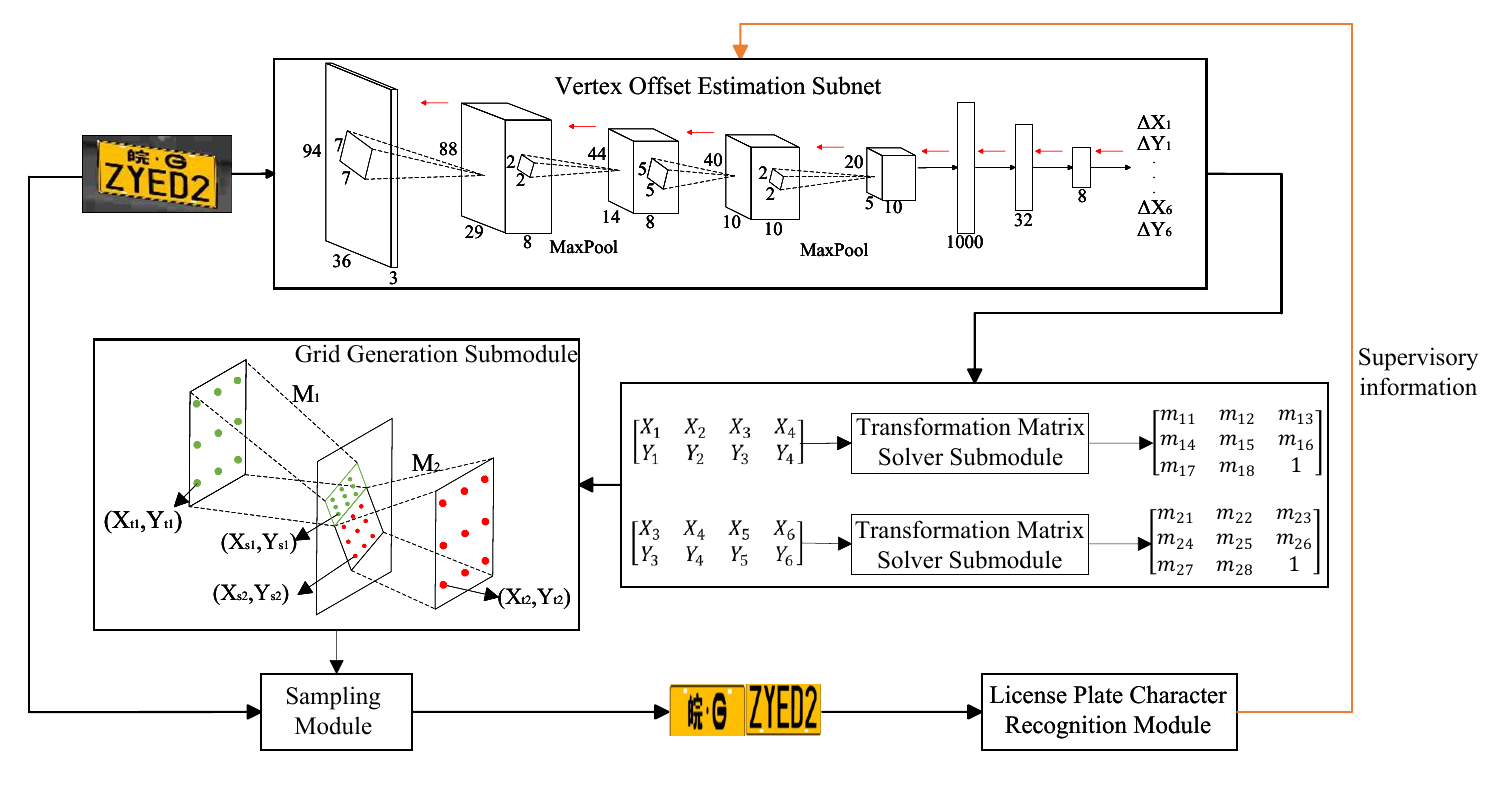}
    \caption{Network structure of perspective distortion rectification module for double-line license plate}\label{fig6}
\end{figure}

\begin{figure}
  \centering
    \includegraphics[width=0.9\linewidth]{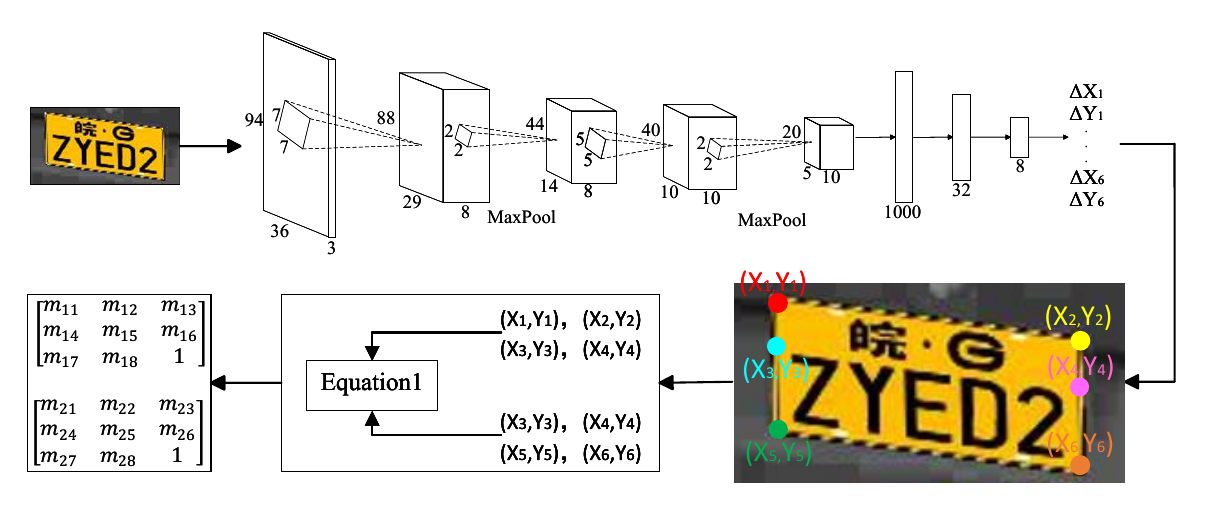}
    \caption{Structure of license plate vertex estimation subnetwork}\label{fig7}
\end{figure}

The correction of double-line license plates is similar to that of single-line license plates, and its network architecture is illustrated in Figure\ref{fig6}. In contrast to the single-line license plate PTR, which only regresses 8 parameters, the vertex offset estimation sub-network for double-line license plates requires regressing 12 parameters. These 12 parameters correspond to the offsets of the vertices of the upper and lower character regions of the double-line license plate relative to the four corner points of the input license plate image. To reduce computational complexity, the upper and lower character regions share two vertices, ultimately yielding 12 offsets, from which 6 vertex coordinates are derived. As depicted in Figure\ref{fig7}, $(x_1,y_1)$ represents the top-left vertex coordinate of the upper character region, $(x_6,y_6)$ represents the bottom-right vertex coordinate of the lower character region, while$(x_3,y_3 )$ and$(x_4,y_4)$ represent the two shared vertex coordinates. Utilizing these 6 vertices, the double-line license plate is segmented into upper and lower parts. Following the same approach as the single-line license plate PTR, perspective transformation matrices are computed for each part. Subsequently, based on these two transformation matrices and in a manner analogous to the single-line license plate PTR, the transformed images corresponding to the upper and lower character regions are obtained. Finally, these two transformed images are concatenated horizontally and fed into a license plate recognition model, leveraging the single-line license plate recognition network to supervise the training of the double-line license plate correction network.
\subsection{Optimized LPRNet with lightweight per-channel attention and Focal Loss}\label{subsec:Optimized-LPRNet-with-lightweight-per-channel-attention-and-Focal-Loss}
\begin{figure}
  \centering
    \includegraphics[width=0.9\linewidth]{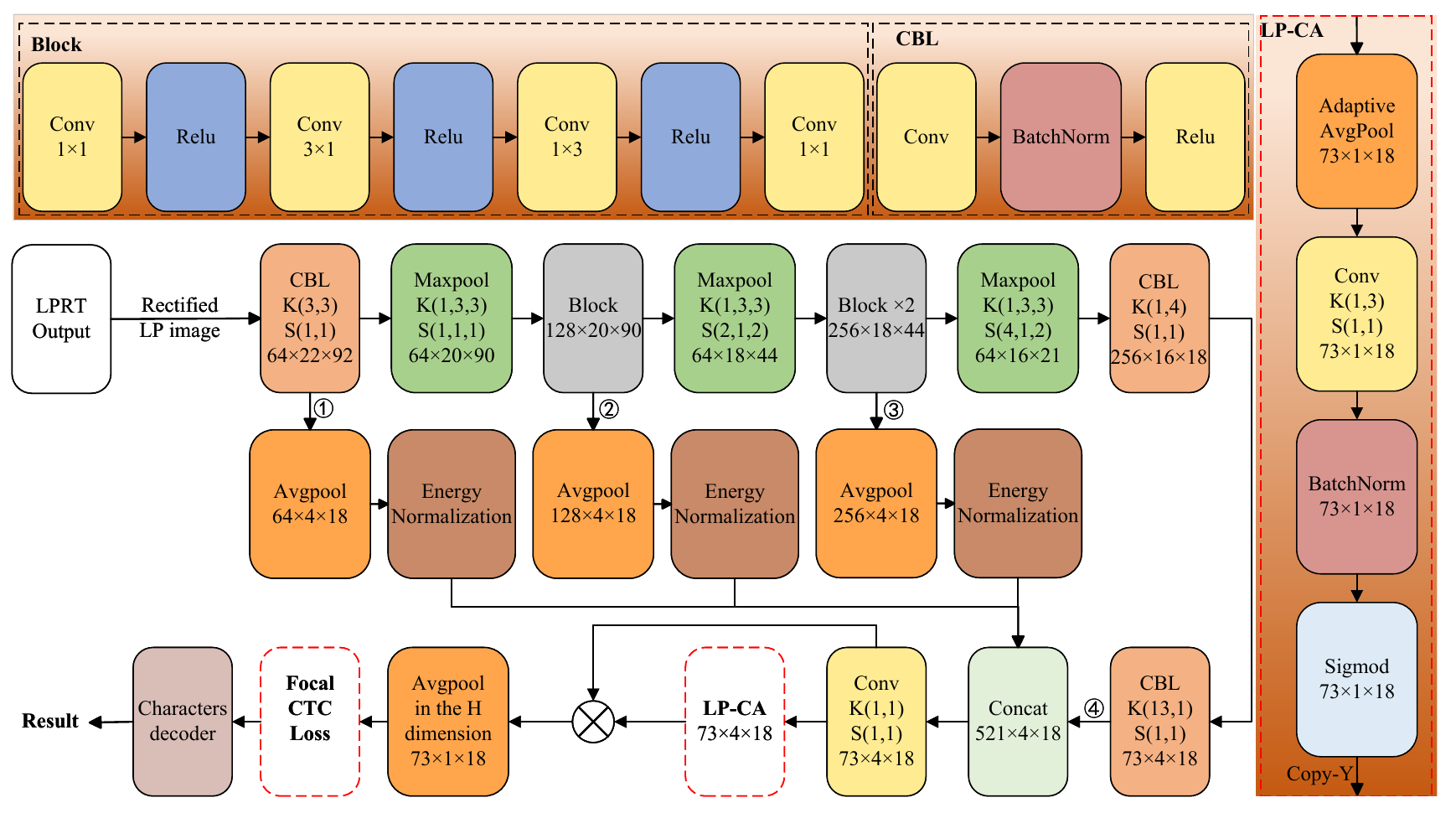}
    \caption{Recognition model network structure diagram}\label{fig8}
\end{figure}

To enhance the accuracy of the recognition network, we introduce a lightweight per-channel attention (LP-CA) module. This module aims to improve the original LPRNet's ability to distinguish between easily confused characters. Furthermore, Focal CTC Loss is employed to mitigate the issue of character class imbalance during training. The architecture of the recognition network is illustrated in Figure\ref{fig8}, with the red dashed box highlighting the modifications introduced in this work.
\subsubsection{LPRNet Optimization with LP-CA}\label{subsubsec:LPRNet-Optimization-with-LP-CA}
\begin{figure}
  \centering
    \includegraphics[width=0.7\linewidth]{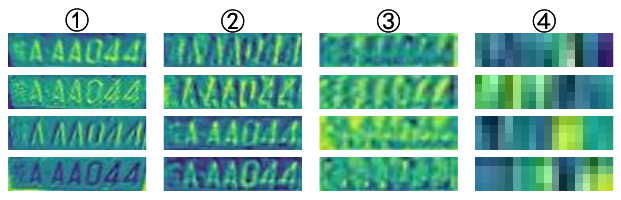}
    \caption{Layer-wise Feature Visualization}\label{fig9}
\end{figure}

As shown in Figure\ref{fig8}, LPRNet employs a multi-level feature fusion architecture, as shown in Figure\ref{fig9}, high-level features focus more on global semantic information, while mid- and low-level features pay more attention to local details.. Specifically, high-level (indicated by \ding{172} in Figure\ref{fig8}), sub-high-level (\ding{173} in Figure\ref{fig8}), mid-level (\ding{174} in Figure\ref{fig8}), and low-level (\ding{175} in Figure\ref{fig8}) features are energy-normalized and then concatenated along the channel dimension to form a comprehensive feature map of size 521 × 4 × 18. Subsequently, 73 convolutional kernels of size 521 × 1 × 1 act as character recognizers, performing point-wise comprehensive interpretation on the 4 × 18 = 72 feature vectors, each with 521 channels. This process yields 73 "lexicon pages" of size 4 × 18, with each page corresponding to a specific character.

During training, guided by the loss function and the predefined ordering of the lexicon, the 73 lexicon pages corresponding to individual character types are gradually aligned with the character library. To obtain the final output for each character position, LPRNet performs column-wise average pooling on the 4×18-sized lexicon pages, resulting in a 1×18 feature vector. This column-wise pooling strategy effectively mitigates the impact of vertical positional variations of the characters on the license plate. Additionally, it helps reduce model parameters and computational complexity by avoiding the need for stacking more convolutional layers to enlarge the receptive field. However, this approach has limitations: each character classifier can only focus on a single position within each column, lacking global context. If the four feature vectors in each column are combined for a more comprehensive representation, the number of parameters would increase fourfold, and the vertical fluctuations of character positions would further complicate feature integration, potentially degrading recognition performance.
\begin{figure}
  \centering
    \includegraphics[width=0.7\linewidth]{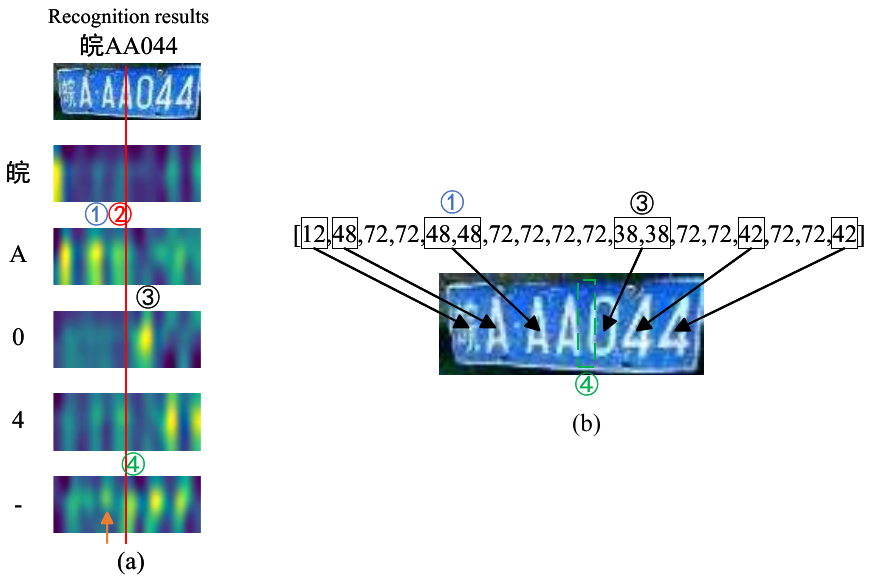}
    \caption{Visualization results of dictionary page feature map and CTC output character ID (a) Visualization results of dictionary page feature map (b) Visualization results of CTC output character ID}\label{fig10}
\end{figure}
As depicted in Figure\ref{fig10}(a), this display visualizes a 4x18 lexicon page feature map generated by LPRNet from a license plate image. Only five lexicon page results are shown here; notably, position \ding{173} indicates an unrecognized 'A' character. Figure\ref{fig10} (b) presents the character IDs output by the CTC decoder, where '48' corresponds to 'A', '38' to '0', and '72' to '-', which serves as an inter-character separator, as highlighted by the green dashed box \ding{175}.

As indicated by the red line in  Figure\ref{fig10} (a), the third 'A' character appears to be influenced by several factors, leading it to resemble '4' more closely than the other two 'A'. Consequently, its activation response is weaker in the dictionary page associated with 'A', but stronger in the page for '4' compared to the other two 'A's. Due to multiple unpadded convolution and pooling operations during feature extraction, it's challenging to perfectly align the character positions in the input image with the 4×18 vector visualization produced by LPRNet, especially near the left and right edges of the image. Furthermore, these repeated convolutional and pooling operations also introduce a degree of mutual interference between characters.

For instance, the response of the character spacing region marked as \ding{175} in Figure (a) is stronger than that between the second 'A' and the third 'A'. Conversely, the spacing region pointed to by the orange arrow, being situated between two consecutive 'A' characters, is more susceptible to the influence of the strong 'A' features when multi-level convolutions extract local patterns. As a result, the activation value in its corresponding dictionary page (marked with '-' in the figure) is somewhat weaker compared to the location indicated by the green \ding{175} position. At \ding{175}, however, because the characters on its left ('A') and right ('0') are distinct, they don't generate a cooperative enhancement effect in their respective feature channels. This leads to a relatively higher response in the character spacing region at this position. Nevertheless, due to the weaker activation of the 'A' character at the red-lined position, combined with the comparatively stronger activation of the adjacent spacing area, the activation value for '-' at this spot slightly exceeds that of 'A'. This ultimately causes the 'A' here to be misidentified as a spacing character ('-'), leading to the loss of this 'A' character during the decoding process.

\begin{figure}
  \centering
    \includegraphics[width=0.6\linewidth]{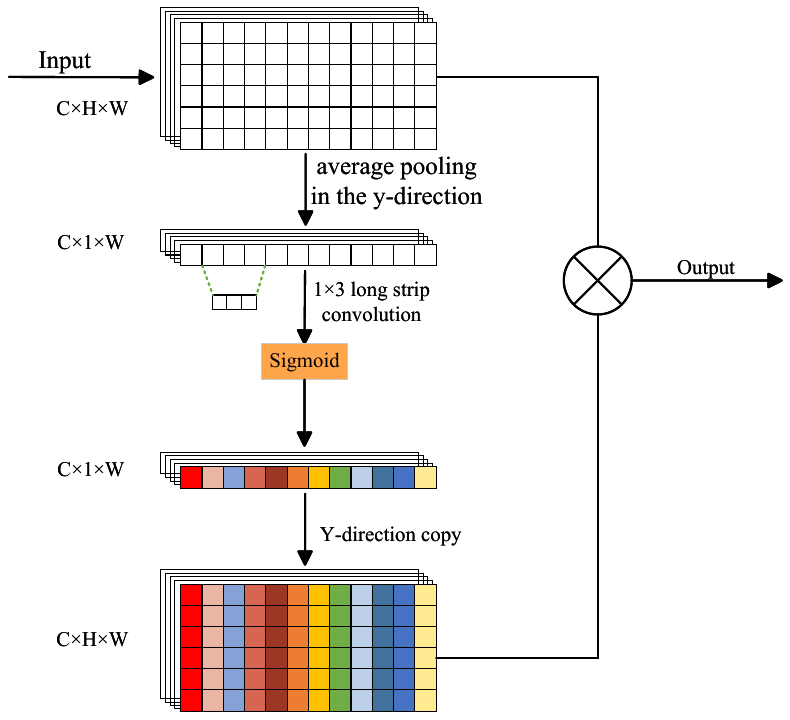}
    \caption{LP-CA network structure}\label{fig11}
\end{figure}

To address the aforementioned issues, this paper proposes a Lightweight Per-channel Attention (LP-CA) module. As illustrated in Figure\ref{fig11}, this module is designed to enhance the high-level features within the network, specifically the 73×4×18 features shown in Figure\ref{fig8}.

In our design, we employ a channel-wise separation strategy. To further reduce computational complexity, we first apply average pooling along the column direction for each channel, yielding a 1×18 feature vector. Subsequently, a 1×3 convolutional kernel is independently applied to each channel to compute its corresponding attention value. The convolutional kernels for each channel are distinct. Through the training process, the network autonomously learns the most suitable kernel parameters for each channel, tailored to the characteristics of the high-level features extracted by that specific channel. This enables effective enhancement of different feature channels.

Once the attention weights are obtained, they are vertically expanded and replicated to form a 73×4×18 attention map (with values ranging from 0 to 1). This map is then element-wise multiplied with the original features to achieve feature reinforcement. The detailed computation is presented in Equation \ref{equation5}.
\begin{equation}
Y=X\otimes \mathrm{Copy}_y (Sig(Conv_{1\times 3} (AP_y (X))))\label{equation5}
\end{equation}

Where $X\in R^{C\times H\times W}$denotes the input channel feature map, and Y represents the resulting enhanced feature map. $AP_y$ signifies the average pooling operation performed along the y-axis (column direction).$Conv_{1\times 3}$ denotes the convolution operation employing a 1×3 convolutional kernel. Sig represents the Sigmoid function, which is utilized to derive attention values within the range of 0 to 1. $Copy_y$ indicates the copy and expansion operation along the y-axis. Finally, $\otimes$ symbolizes element-wise multiplication.
\subsubsection{LPRNet Optimization Based on Focal CTC Loss}\label{subsubsec:LPRNet-Optimization-Based-on-Focal-CTC-Loss}
\begin{figure}
  \centering
    \includegraphics[width=0.9\linewidth]{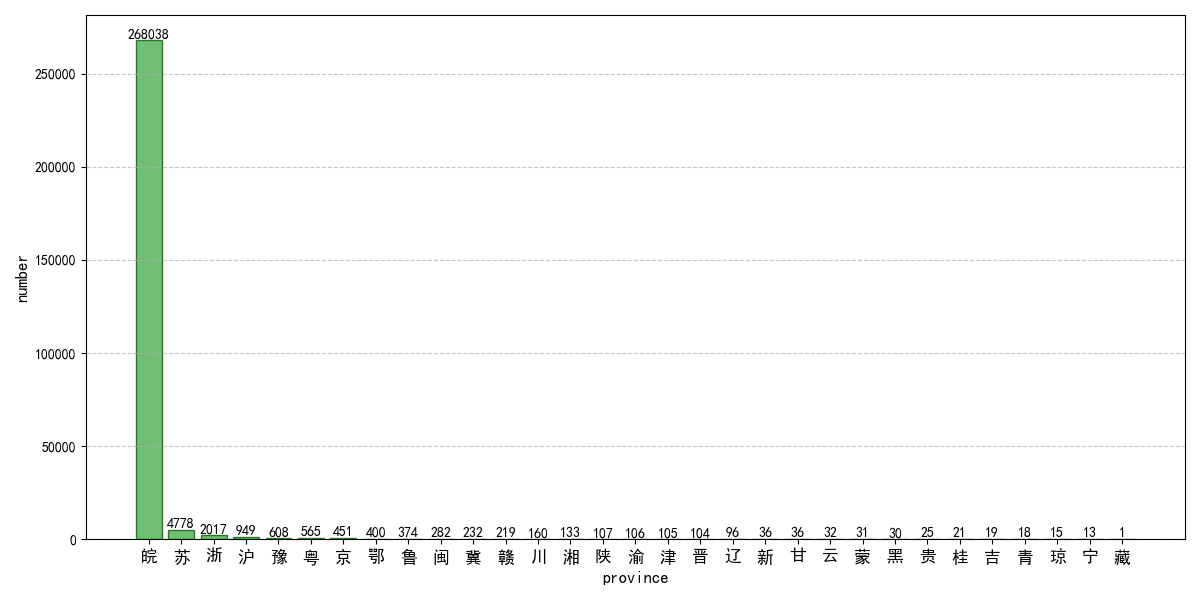}
    \caption{CCPD License Plate Distribution by Province}\label{fig12}
\end{figure}
The CCPD license plate dataset exhibits a significant class imbalance in the distribution of Chinese characters. Statistical analysis of the CCPD dataset (Figure\ref{fig12}) reveals that license plates beginning with the character "WAN" constitute 95.7\% of the dataset. Furthermore, the overall frequency of Chinese characters within the entire character set is only 14.29\%, considerably lower than that of alphanumeric characters.

The existing Connectionist Temporal Classification (CTC) loss function\cite{graves2006connectionist}fails to effectively address the issue of data imbalance in character distribution. During the loss computation, it assigns equal weights to all possible output paths, neglecting the varying frequencies of different characters. Consequently, high-frequency characters dominate the training process, resulting in insufficient learning of low-frequency Chinese characters. This often leads to misclassification of the initial Chinese character in license plates from provinces containing a small number of low-frequency characters, with a common example being the misrecognition of the first character as "WAN".

Focal Loss\cite{lin2017focal} is a loss function specifically designed to address the problem of severe class imbalance. Its expression is given by Equation \ref{equation6}:
\begin{equation}
\left\{
\begin{array}{l l}  
    -\alpha(1-\alpha)^\gamma \log p, & \text{Positive sample} \\
    -(1-\alpha)p^\gamma \log(1-p),  & \text{Negative sample} \\
\end{array}
\right.\label{equation6}
\end{equation}

Where $p$ is the predicted probability of the sample; $\alpha$ is the balancing factor used to mitigate class imbalance between positive and negative samples; and $\gamma$ is the adjustment factor designed to modulate the rate at which the weights of easy samples are reduced, thereby encouraging the network to focus more on hard samples.

In the context of license plate recognition tasks, traditional negative samples are absent. To address this, we adapt the standard Focal Loss approach for positive samples in the calculation of our Focal CTC Loss. This method aims to mitigate the disproportionate contribution of easily classified samples, thereby focusing training on more challenging instances. Implementation-wise, this loss replaces the conventional CTC Loss used in the original LPRNet. During CTC decoding, a greedy search is performed over all characters in the license plate to derive the probability value for each character. By multiplying these probabilities, we obtain $p$, the total probability of correctly recognizing the entire license plate image. Subsequently, we introduce a weighting factor based on $p$ to compute the Focal CTC Loss. The calculation formula is as follows:
\begin{equation}
\text{Focal CTC loss}=\alpha(1-p)^\gamma*\text{CTC Loss}
\end{equation}

Focal CTC Loss enables license plate recognition networks to focus on challenging samples, thereby mitigating the adverse effects of class imbalance. In our experiments, we employ a balance factor $\alpha$=0.5and an adjustment factor $\gamma$=2.0. The results demonstrate that the Focal CTC Loss trained model achieves a 17\% improvement in accuracy on license plates containing infrequent Chinese characters, and a 0.42\% improvement in overall license plate recognition accuracy, compared to the model trained with standard CTC Loss. These findings suggest that Focal CTC Loss effectively alleviates the class imbalance problem and enhances the robustness of the model.
\subsection{Double-line License Plate Dataset Construction}\label{Double-line-License-Plate-Dataset-Construction}
\begin{figure}
  \centering
    \includegraphics[width=0.9\linewidth]{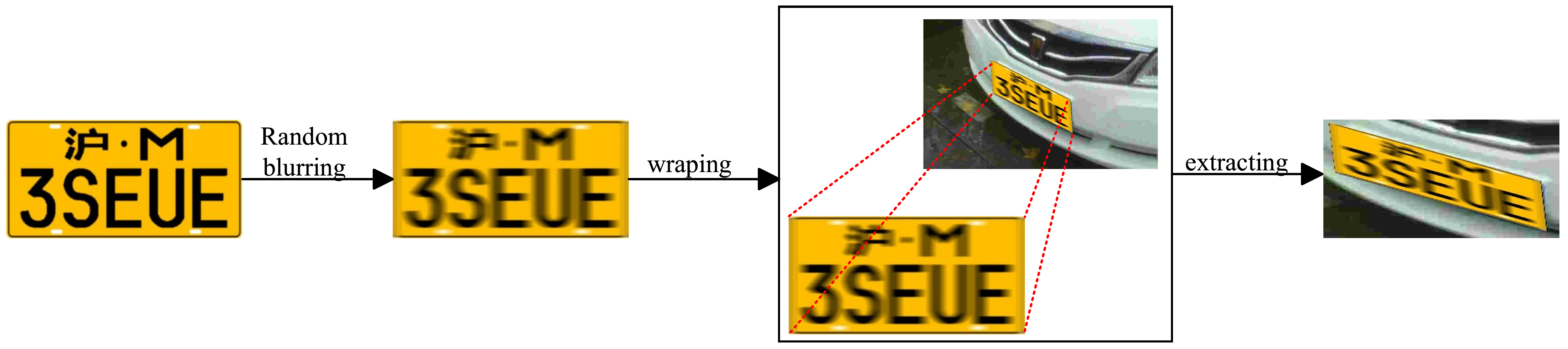}
    \caption{Main flowchart for building a double-line dataset}\label{fig13}
\end{figure}

\begin{figure}
  \centering
    \includegraphics[width=0.7\linewidth]{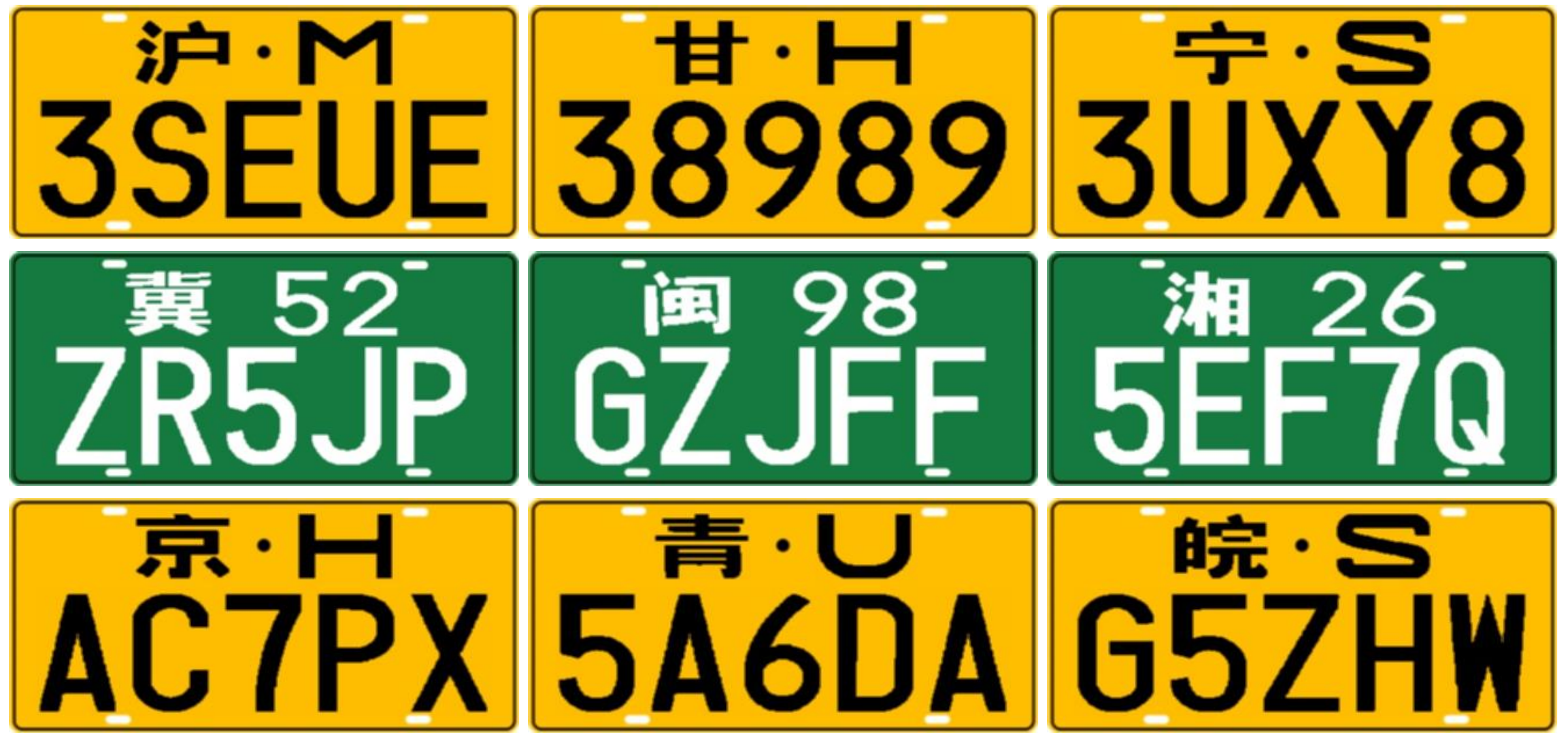}
    \caption{Example of a generated double-line license plate}\label{fig14}
\end{figure}

\begin{figure}
  \centering
    \includegraphics[width=0.7\linewidth]{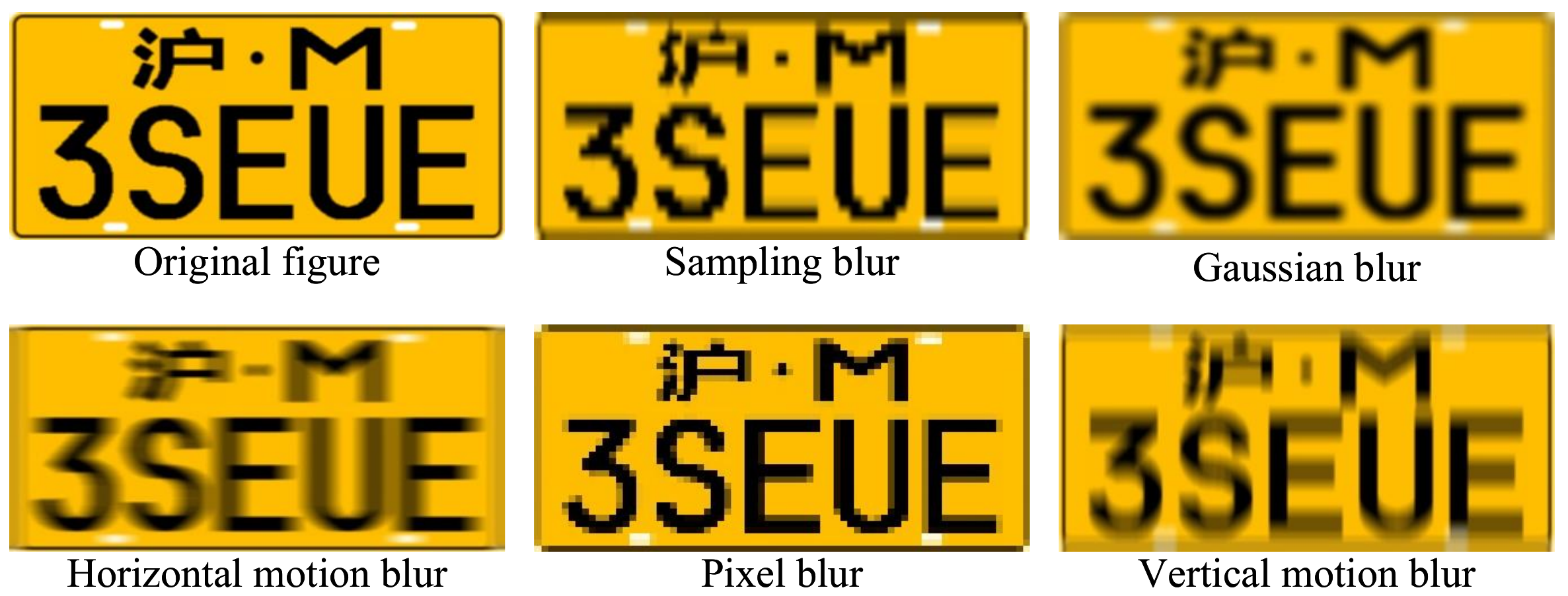}
    \caption{Various blurring effects}\label{fig15}
\end{figure}

Currently, research in the field of license plate recognition (LPR) predominantly focuses on single-line license plates. Studies targeting double-line license plates remain relatively limited, primarily due to the scarcity of publicly available, large-scale datasets of such plates. To address this gap, this paper constructs a dedicated double-line license plate dataset. The detailed construction process is illustrated in Figure\ref{fig13}.
The construction of the dataset mainly encompasses the following key steps: First, to ensure the diversity of generated characters, we utilized meticulously designed hand-crafted templates to generate a large number of double-line license plate images (as shown in Figure\ref{fig14}), striving to cover as many character combinations as possible. Second, to enhance the realism of the dataset, we employed random blurring techniques for data augmentation of the generated images, simulating common phenomena in real-world scenarios such as motion blur and defocus blur (as shown in Figure\ref{fig15}), thereby making them closer to actual photographic effects. Finally, to simulate multi-angle perspectives and complex lighting conditions in open environments, we referenced the license plate vertex coordinates provided in the CCPD dataset to apply perspective transformations to the generated double-line license plate images, overlaying the transformed images into corresponding CCPD sample images. 

This approach not only ensures the close alignment of the license plate image's viewing angle with real-world scenes but also integrates it into the background regions of some real scenes, thereby more effectively simulating license plate recognition scenarios in complex environments. This methodology more closely reflects real-world applications and contributes to improved model generalization capabilities.

\begin{figure}
  \centering
    \includegraphics[width=0.5\linewidth]{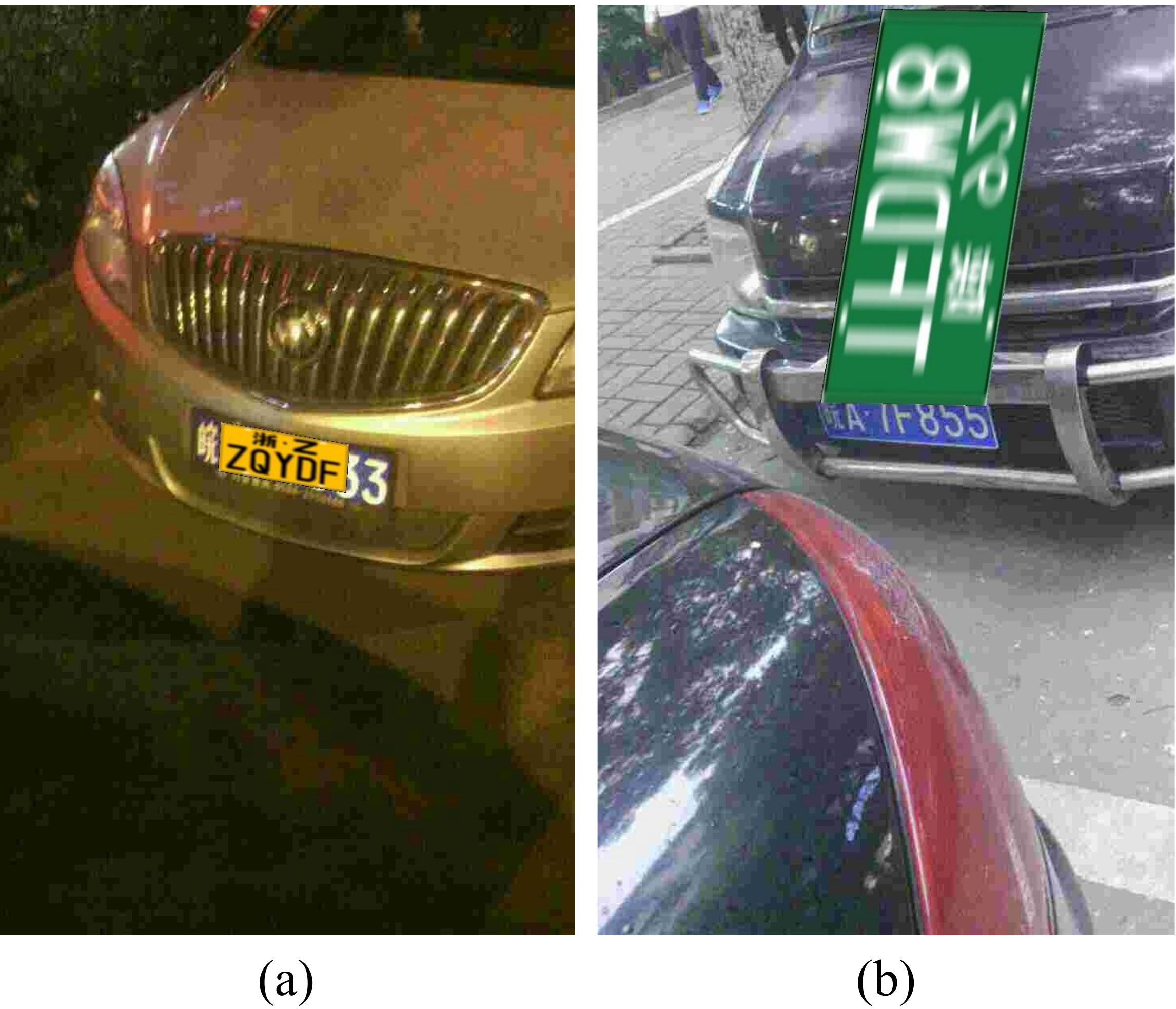}
    \caption{Examples of Labeling Errors in CCPD}\label{fig16}
\end{figure}

\begin{figure}
  \centering
    \includegraphics[width=0.65\linewidth]{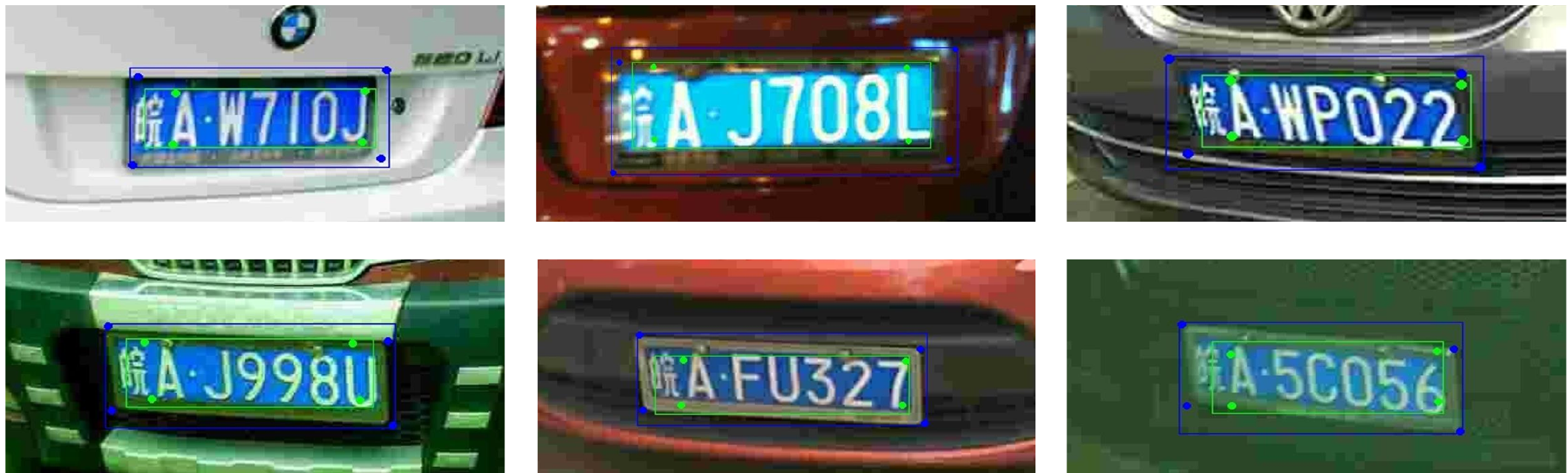}
    \caption{Examples of Selected Mislabeled License Plate Images}\label{fig17}
\end{figure}

\begin{figure}
  \centering
    \includegraphics[width=0.65\linewidth]{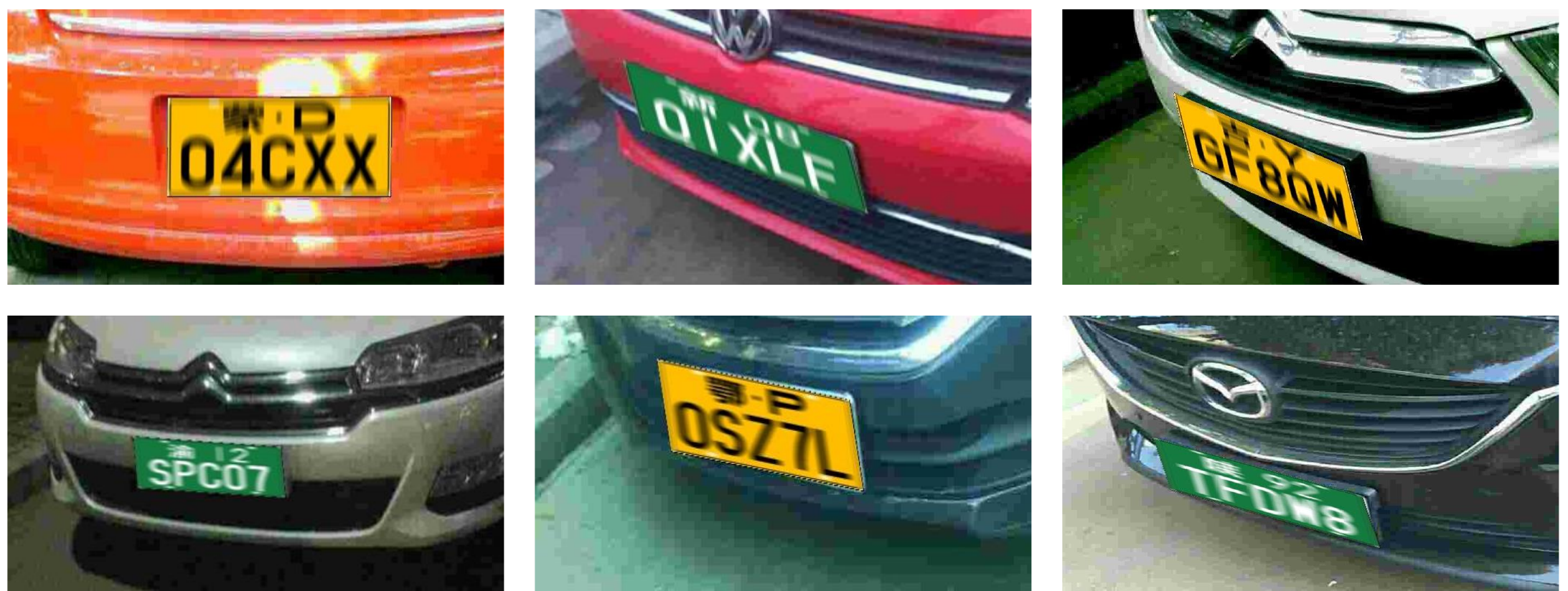}
    \caption{Sample Images from the Two-Line License Plate Dataset}\label{fig18}
\end{figure}

Through examination of the generated dataset of double-line license plates, we identified labeling inaccuracies inherent within the CCPD dataset, resulting in the scenarios illustrated in Figure\ref{fig16}. While instances of errors depicted in Figure\ref{fig16} (b) were infrequent, the majority of cases corresponded to the type shown in Figure\ref{fig16} (a). To address this issue, we implemented a correction procedure for the CCPD labels. Specifically, we first employed a license plate detection model to identify license plate regions. The veracity of each label was then assessed based on the Intersection over Union (IoU) between the predicted bounding box and the ground truth label. An IoU exceeding 0.6 was considered indicative of a correct label; otherwise, the label was deemed erroneous. This screening process revealed a total of 1414 images containing mislabeled license plates (as shown in Figure\ref{fig17}, where blue bounding boxes represent the predicted locations and green boxes denote the original labels). These images with erroneous labels were subsequently subjected to manual annotation for rectification, ensuring label accuracy and, consequently, enhancing the overall quality of the dataset.
Finally, the corrected CCPD dataset was used for image composition to generate the final double-line license plate dataset (as shown in Figure\ref{fig18}). This dataset contains a total of 200,000 license plate images, divided into training, validation and test sets at an 8:1:1 ratio.
\section{Experiment settings and Results Analysis}\label{sec:Experiment-settings-and-Results-Analysis}
\begin{figure}
  \centering
    \includegraphics[width=0.65\linewidth]{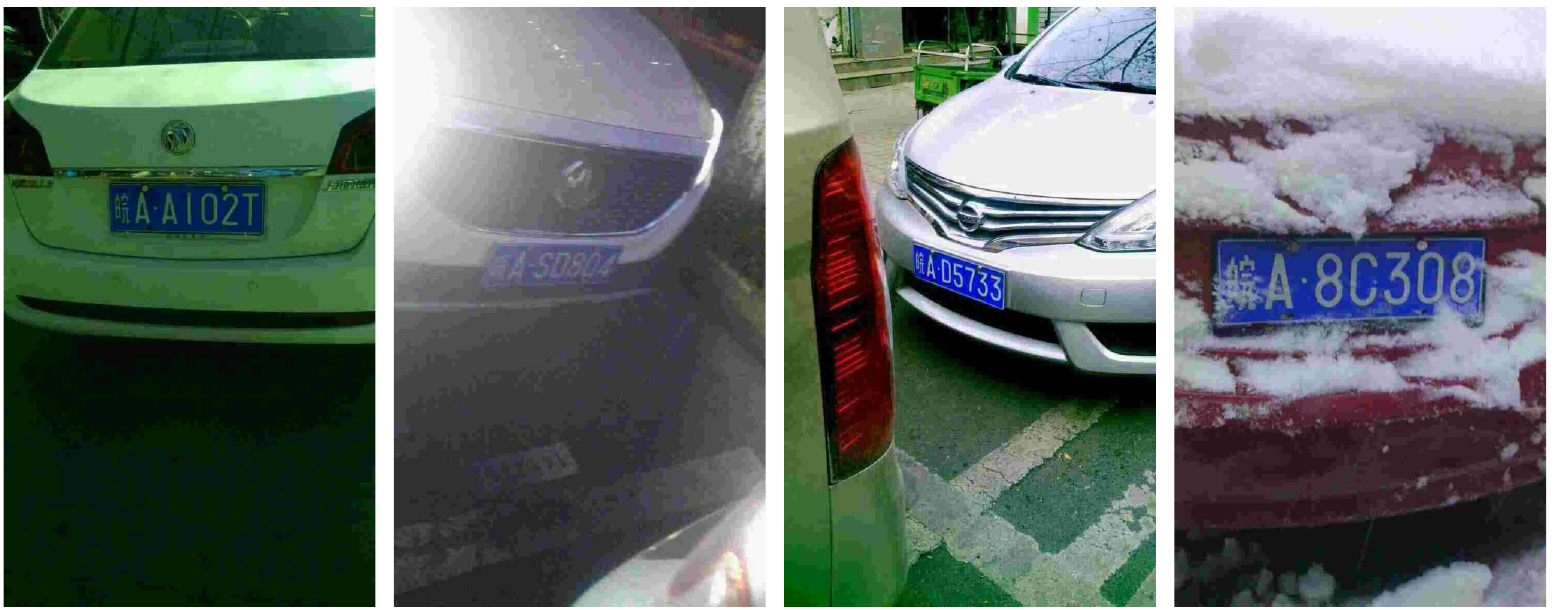}
    \caption{Samples of CCPD dataset}\label{fig19}
\end{figure}

\begin{figure}
  \centering
    \includegraphics[width=0.65\linewidth]{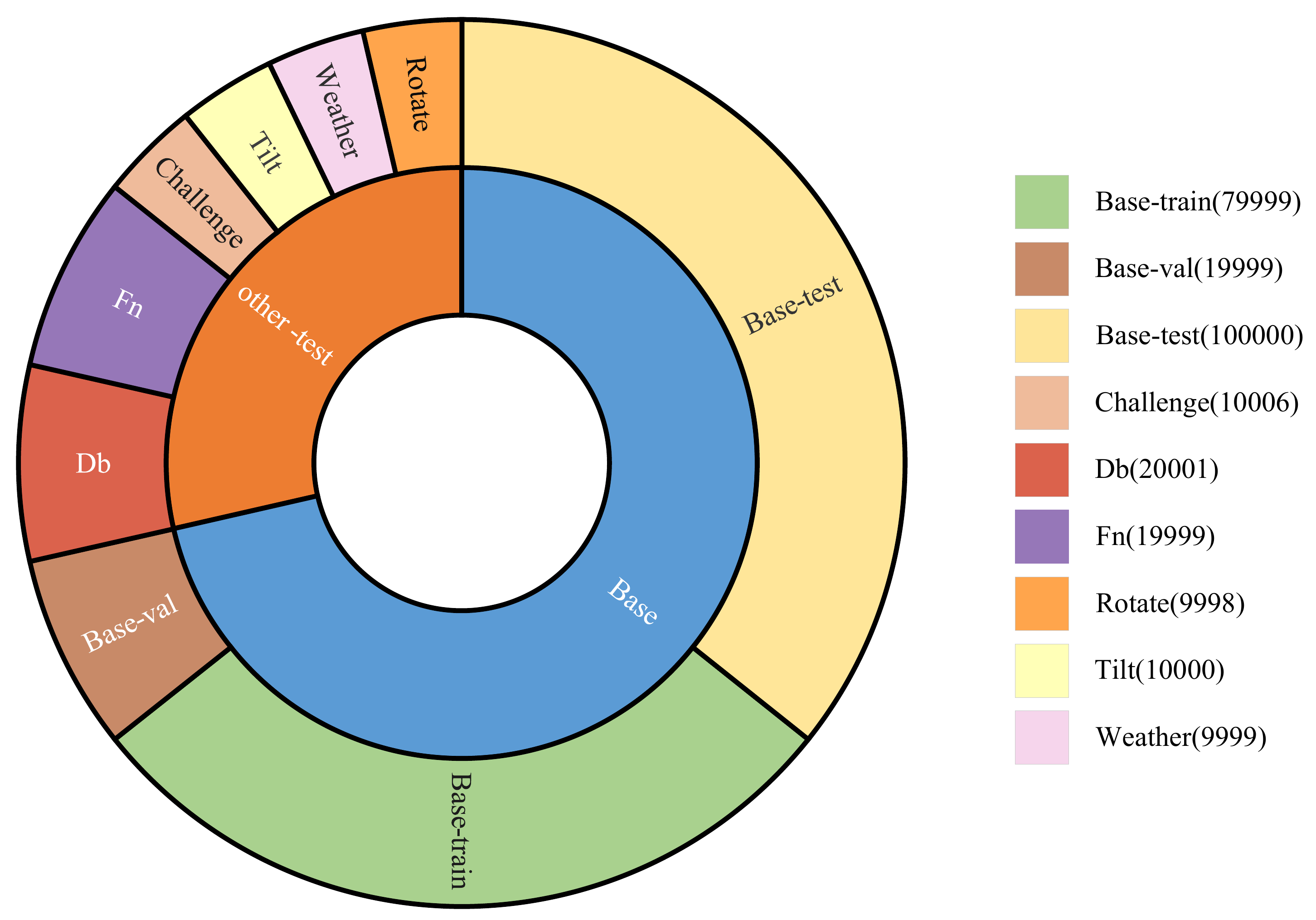}
    \caption{Distribution Map of CCPD Subsets and License Plate Types}\label{fig20}
\end{figure}

\begin{figure}
  \centering
    \includegraphics[width=0.65\linewidth]{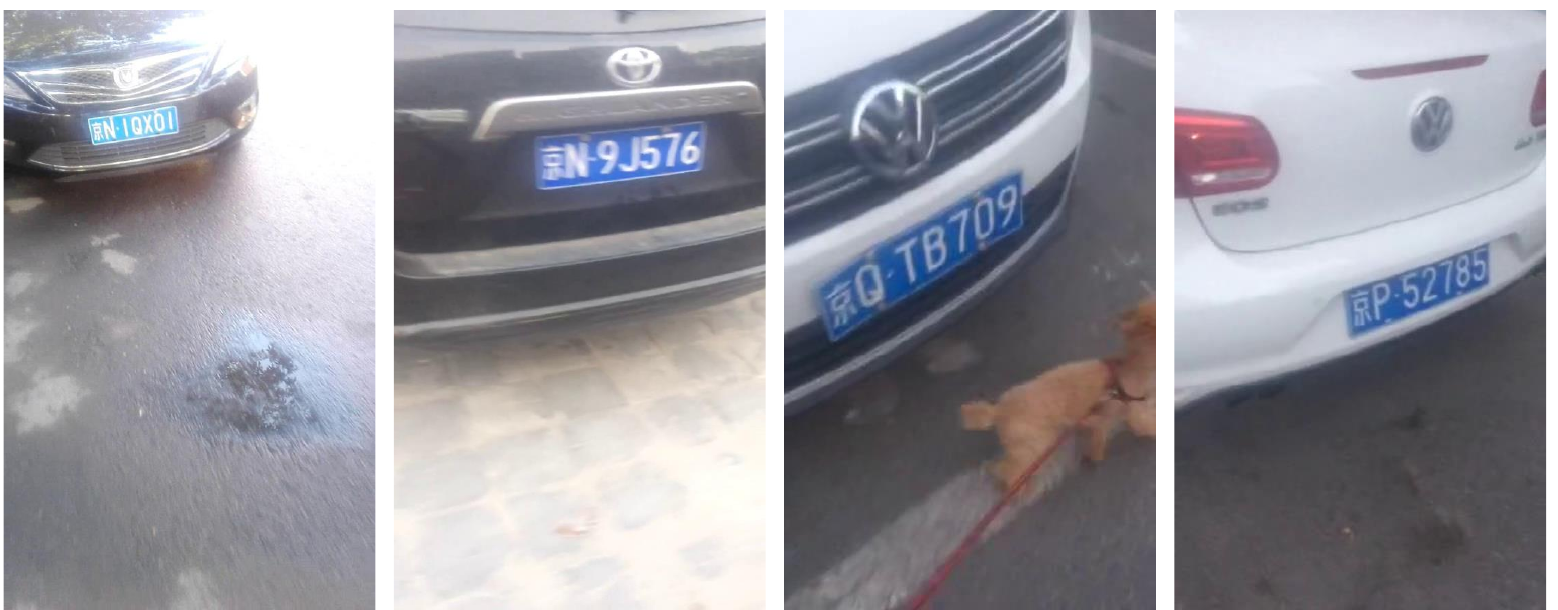}
    \caption{Samples of LSV dataset}\label{fig21}
\end{figure}

\begin{figure}
  \centering
    \includegraphics[width=0.65\linewidth]{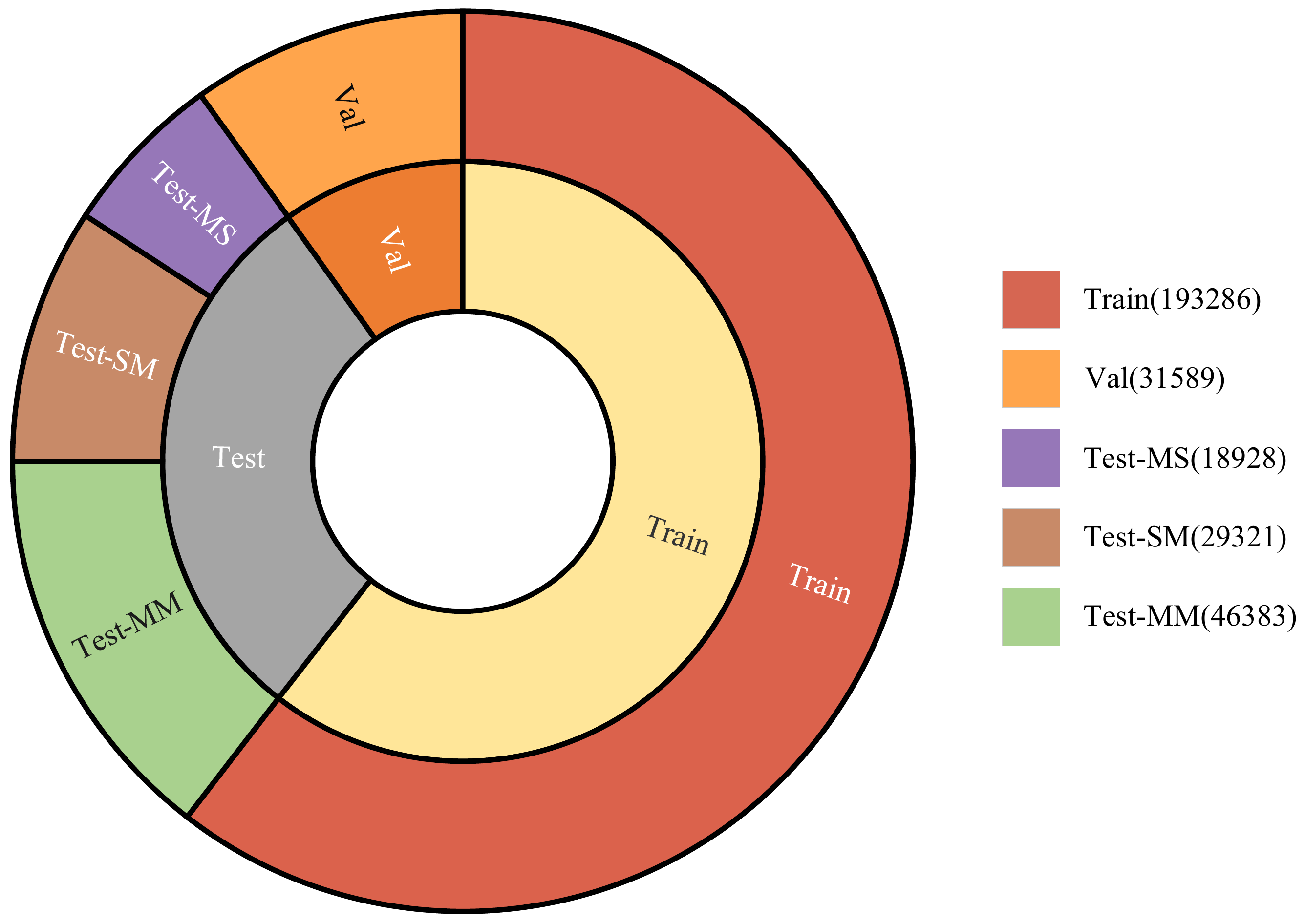}
    \caption{Distribution Map of LSV Dataset}\label{fig22}
\end{figure}

The dataset used in this experiment is the publicly available CCPD license plate dataset. This dataset encompasses license plate images captured in complex environmental conditions, including various challenging scenarios such as illumination variations, tilting, rain, and snow (as shown in Figure\ref{fig19}), thereby providing a comprehensive basis for evaluating the performance of recognition algorithms. The primary goal of this study is to validate the effectiveness of the proposed algorithm using the CCPD dataset. Regarding data partitioning, we adhered to the official CCPD recommended split to ensure consistency with previous literature. Specifically, the Base subset was randomly divided into two equally sized parts, with one part designated as the final testing set. The other part was further split into training and validation sets at an 80\%/20\% ratio. The remaining subsets of the CCPD dataset were incorporated into the final testing set. The overall data partitioning scheme is illustrated in Figure\ref{fig20}.
The model was also evaluated using the LSV dataset\cite{wang2022lsv} which contains static and dynamic characteristics. The LSV dataset categorizes samples into three types based on the relative motion between photographer and vehicle: static vs move, move vs static, and move vs move (as shown in Figure\ref{fig21}). Since some LSV images lack license plates, we excluded plate-free images and retained only plate-containing samples for training. The dataset partitioning strictly follows LSV's original scheme with train, valid, and test subsets, with specific division details shown in Figure\ref{fig22} to ensure experimental fairness.
\subsection{License plate localization error simulation based on random perturbation}\label{subsec:License-plate-localization-error-simulation-based-on-random-perturbation}
\begin{figure}
  \centering
    \includegraphics[width=0.8\linewidth]{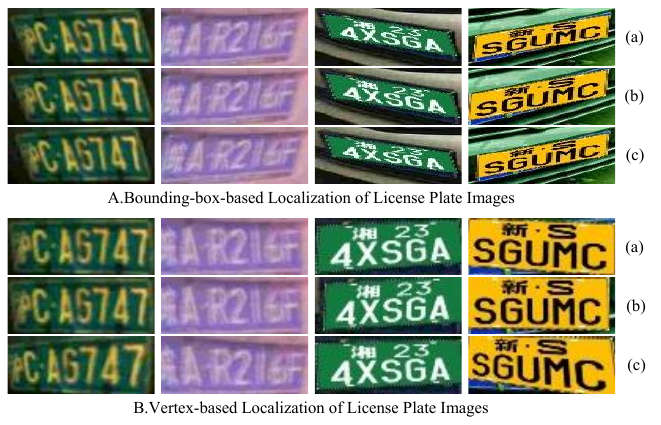}
    \caption{License plate images obtained using different extraction methods: (a) Images based on ground truth labels, (b) Images based on YOLO detection results, (c) Images based on random coordinate perturbation of ground truth labels}\label{fig23}
\end{figure}

Accurate simulation of localization errors is crucial for training robust license plate recognition models, given the independent nature of license plate localization and recognition. To better mimic the errors encountered during real-world localization and decouple these tasks, this paper employs a random perturbation method to simulate license plate detection labels. This process aims to generate more realistic license plate images, providing the recognition network with representative training samples. Specifically, Gaussian noise (mean=0, std=4 pixels) is added independently to the top-left and bottom-right corner coordinates of the license plate bounding box in the original labels to randomly perturb these positions. Additionally, the coordinates of each of the four vertices of the license plate are also randomly perturbed to further simulate vertex localization errors. This perturbation introduces localization errors into the training labels, significantly increasing training sample diversity. Figure\ref{fig23} illustrates a comparison of license plate images before and after perturbation: Figure\ref{fig23} (a) shows the original labels from the CCPD dataset; Figure\ref{fig23} (b) shows the bounding box and vertex coordinate results obtained from a YOLO detection model\cite{xu2022accurate}; and Figure\ref{fig23} (c) shows the license plate image generated after applying the random coordinate perturbation. This error simulation effectively reflects real-world localization errors, enhancing the robustness of the recognition model in complex scenarios. Consequently, the proposed perturbation method provides favorable conditions for evaluating license plate recognition networks independently of localization networks.
\subsection{LPTR-AFLNet Training}\label{subsec:LPTR-AFLNet-Training}
The software platform for this experiment is based on Ubuntu 18.04, PyTorch 1.8, and CUDA 11.1. The hardware environment comprises a training machine and several evaluation/testing machines. The training machine features a TITAN X GPU, an Intel Xeon E5-2620 CPU, and 80 GB of memory. Evaluation machine 1 has an identical configuration to the training machine. Evaluation machine 2 is equipped with an Intel Core i7-7700K CPU and 24 GB of memory, and evaluation machine 3 is equipped with an Intel Core i7-4770K CPU and 32 GB of memory. The hyperparameter settings used for network training are detailed in Table\ref{tbl1}.

\begin{figure}
  \centering
    \includegraphics[width=0.8\linewidth]{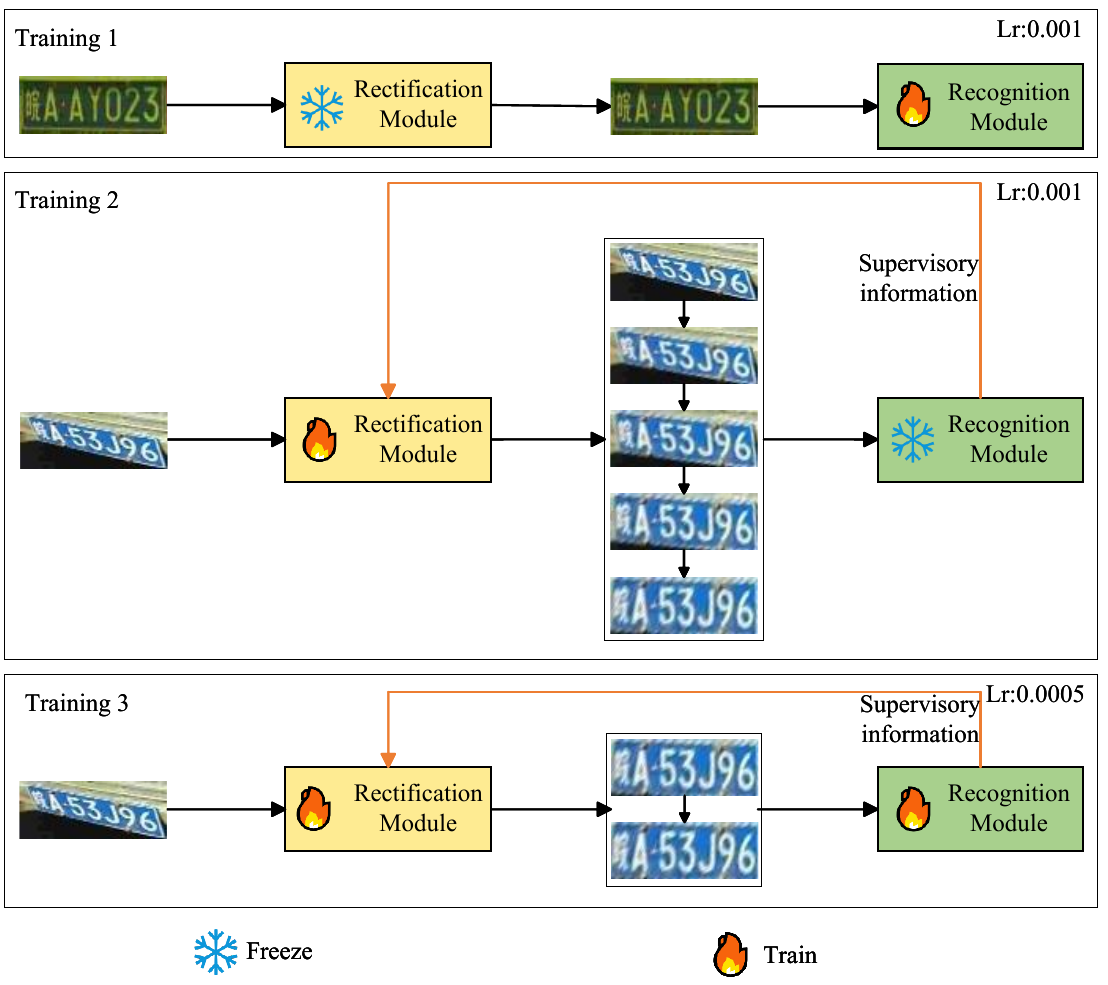}
    \caption{Illustration of LPRTR-AFLNet Training Strategy}\label{fig24}
\end{figure}

To enhance the convergence speed and overall performance of the weakly supervised, recognition-driven, unconstrained Chinese license plate automatic rectification algorithm during training, a staged training strategy as illustrated in Figure\ref{fig24} has been designed. The approach begins by freezing certain weights of the plate rectification network and utilizing precisely localized license plate images obtained through vertex localization to pretrain the LPRNet recognition network, thereby establishing initial recognition weights. Subsequently, the pretrained recognition network weights are used as a fixed initialization, with the LPRNet network itself frozen, and only the rectification network is trained using license plate images detected by rectangular bounding boxes. This step aims to gradually guide the rectification network in learning the correction task while leveraging the auxiliary guidance provided by the weakly supervised recognition information. Finally, based on the previous training stages, both the rectification network and the recognition network are jointly fine-tuned to optimize the overall model parameters, achieving improved training efficiency and optimal performance.
\begin{table}
\caption{Model training parameters}\label{tbl1}
\begin{tabular*}{\tblwidth}{@{}CC@{}}
\toprule
Parameters& Numerical parameters  \\ 
\midrule
Batch Size& 300\\
Optimizer& Adam\\
Learning Rate&0.001 and 0.0005 \\
Momentum &0.9 \\
Epoch&200 \\
\bottomrule
\end{tabular*}
\end{table}
\subsection{Experimental results analysis}\label{subsec:Experimental-results-analysis}
\subsubsection{PTR performance analysis}\label{subsubsec:PTR-performance-analysis}
\begin{figure}
  \centering
    \includegraphics[width=0.8\linewidth]{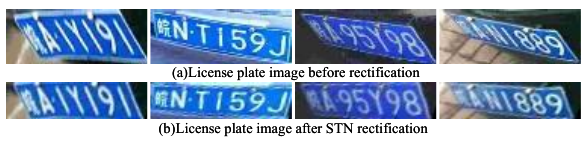}
    \caption{Samples of affine transformation rectification with STN}\label{fig25}
\end{figure}

\begin{figure}
  \centering
    \includegraphics[width=0.8\linewidth]{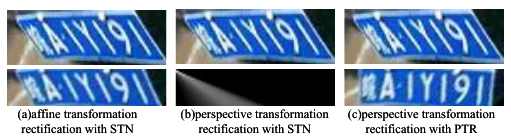}
    \caption{Comparison of rectification results of STN and PTR}\label{fig26}
\end{figure}

\begin{figure}
  \centering
    \includegraphics[width=0.8\linewidth]{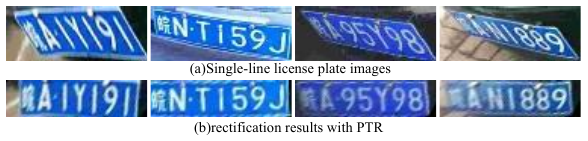}
    \caption{Single-line license plate image rectification samples with PTR}\label{fig27}
\end{figure}

Traditional Spatial Transformer Networks (STN) typically regress only the affine transformation matrix, thereby limiting their capability to perform simple planar transformations such as cropping, rotation, and scaling. Consequently, their effectiveness in correcting license plate images with perspective distortion is limited, as illustrated in Figure\ref{fig25}. Theoretically, STN can be extended to estimate a variety of complex spatial transformation parameters for more accurate correction of target regions. In this study, we attempted to directly regress perspective transformation parameters using STN; however, experimental results indicate that, in situations where license plate images exhibit perspective distortions and other complex deformations, STN struggles to effectively regress the perspective transformation parameters. This often leads to poor convergence and outputs chaotic images. As shown in Figure\ref{fig26}, subfigure (a) displays the correction results obtained by STN regressing affine transformation parameters, subfigure (b) shows the results when STN regresses perspective transformation parameters, and subfigure (c) presents the correction results when PTR regresses perspective transformation parameters. It can be observed that, when using STN to regress perspective transformation parameters, the corrected license plate images appear notably chaotic. This is due to the fact that perspective transformations are inherently more complex than affine transformations, with parameters that are interdependent. When using license plate recognition results as weak supervision signals to train STN, these intricate parameter interactions often hinder convergence, resulting in distorted correction outputs.
Following processing with PTR, individual license plate images exhibiting perspective distortion can be effectively rectified, as demonstrated in Figure\ref{fig27}. PTR successfully addresses the performance limitations of STN when correcting license plates with perspective deformation. While the rectified license plates may still exhibit slight tilting, this is attributable to minor inaccuracies in the labels of the CCPD dataset itself and subtle biases in the training dataset, which is considered a normal occurrence. Moreover, the PTN (Perspective Transformation Network) and the recognition network in this work are jointly trained, necessitating only sufficient rectification for accurate recognition. 

\begin{figure}
  \centering
    \includegraphics[width=0.8\linewidth]{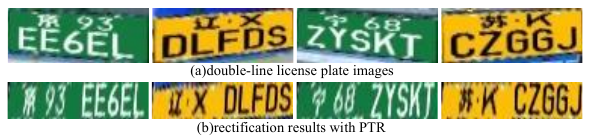}
    \caption{Double-line license plate image rectification samples with PTR}\label{fig28}
\end{figure}

For double-line license plates, the rectification effect of PTR is also remarkably evident, as depicted in Figure\ref{fig28}. After PTR correction, the upper and lower lines of the double-line license plate are simultaneously straightened and horizontally concatenated into a single line, enabling license plate recognition in a single-line format and effectively improving the recognition performance of double-line license plates.
\subsubsection{AFLNet Performance Analysis}\label{subsubsec:AFLNet-Performance-Analysis}
In this study, we conducted comparative experiments on the optimized model and mainstream license plate recognition methods using the CCPD dataset. The comparison results are presented in Table\ref{tbl2} (testing device: Test Machine 1). The upper part of Table\ref{tbl2} displays models with publicly available code. To ensure a fair comparison, these models were all reproduced under our specific hardware and software environment and dataset conditions. The lower part of Table\ref{tbl2} lists the results of integrated detection and recognition models with publicly unavailable code; these models perform recognition based on the results of vertex localization (i.e., only recognizing detected license plates). In contrast, we simulate detection results by applying random perturbations to the vertex localization results and perform recognition on the entire sample set of the CCPD dataset.

As can be seen from Table\ref{tbl2}, our algorithm achieves superior recognition accuracy compared to LPRNet across all subsets of the CCPD dataset, especially with notably significant improvements in the rotate, tilt, and challenge subsets. This is attributed to the PTR module, which can adaptively rectify license plate images to meet the needs of recognition. Nevertheless, our algorithm's accuracy is still slightly lower than that of some heavyweight models\cite{peng2023end}\cite{10458015}. This is due to the fact that those models stack more convolutional layers, resulting in larger network architectures and stronger feature extraction capabilities. However, this strong capability also leads to a decrease in processing speed, making it difficult to meet the demands of practical applications. It is generally considered that a system needs to process at least 30 frames per second to be considered a "real-time" system \cite{laroca1909efficient}. Although the model proposed in this paper compromises somewhat in terms of accuracy, it maintains relatively high recognition performance while achieving speeds far exceeding heavyweight models, thus possessing greater practical value.
\begin{table}
\caption{Performance Comparisons of Different Methods on CCPD}\label{tbl2}
\begin{threeparttable}
\begin{tabular*}{\tblwidth}{@{}CCCCCCCCCCCC@{}}
\toprule
Modle & AP & Base & Db & Fn & Rotate & Tilt&Weather&Challenge&Double&FPS&Size(M) \\ 
\midrule
RPNet\cite{shi2016end} & 95.64 & 98.42 & 96.82 & 95.04 & 90.33 & 92.62&86.93&83.69&×&61&210 \\
LPRNet\cite{zherzdev2018lprnet} & 98.35 & 99.49 & 98.29 & 98.62 & 98.73 & 98.77&97.04&87.08&×&\underline{129}&\textbf{1.8} \\
Eulpr\cite{qin2020efficient} & 98.64 & 99.54 & 98.84 & 98.78 & \textbf{99.15} & \underline{99.15}&97.79&88.87&\underline{98.73}&63&3.9 \\
\bottomrule
Method\cite{seo2022robust}(2022) & 92.60 & 99.83 & 73.32 & 83.13 & 94.11 & 88.26&97.48&75.83&×&-&- \\
Method\cite{gong2022unified}(2022) & 96.57 & 98.30 & 98.00 & 97.20 & 92.50 & 93.70&90.70&87.90&×&30&640 \\
Method\cite{peng2023end}(2023) &98.70 & \textbf{99.85} & 98.78 & 98.80 & 98.11 & 98.80&\textbf{98.90}&88.80&×&26&- \\
Method\cite{10458015}(2024) & \underline{98.80} & \underline{99.70} & \textbf{99.10} & \underline{99.10} & \underline{98.40} & 98.80&\underline{98.50}&\underline{89.50}&×&\textbf{154}&- \\
LPTR-AFLNet & \textbf{98.87} & 99.69 & \underline{98.99} & \textbf{99.11} & \textbf{99.15} & \textbf{99.19}&97.89&\textbf{90.39}&\textbf{99.37}&107&\underline{2.7} \\
\bottomrule
\end{tabular*}
\begin{tablenotes}   
    \footnotesize               
    \item[1]{"x": indicates unsupported; "-": indicates data not publicly available}
\end{tablenotes}            
    \end{threeparttable}
\end{table}

To further evaluate our lightweight model, we conducted experiments on the LSV dataset\cite{wang2022lsv}, employing the same evaluation metrics as in  \cite{wang2022lsv}. Specifically, we used Accuracy\_7c, which measures the accuracy of recognizing all 7 characters of the license plate (including Chinese characters, numbers, and letters), and Accuracy\_6c, which measures the accuracy of recognizing the non-Chinese character portion (numbers and letters) when Chinese characters are not considered. The results are summarized in Table\ref{tbl3}.

Similar to Table\ref{tbl2}, the top half of Table\ref{tbl3} presents the reproduced results of the models from \cite{wang2022lsv} based on their open-source code, while the bottom half directly cites the results from the original LSV paper \cite{wang2022lsv}, as the original authors did not release their code. When comparing our reproduced LPRNet recognition results with the data from \cite{wang2022lsv}, slight differences exist across different subsets. However, the overall accuracy and computational speed (Table\ref{tbl3} uses the same "ms" unit as in \cite{wang2022lsv}) are largely consistent. These discrepancies may stem from differences in hardware platforms.
\begin{table}
\caption{Performance Comparisons of Different Methods on LSV}\label{tbl3}
\begin{threeparttable}
\begin{tabular*}{\tblwidth}{@{}CCCCCCCCCC@{}}
\toprule
\multirow{2}*{Subsets} & \multicolumn{2}{c}{Static vs Move}  &\multicolumn{2}{c}{Move vs Static} & \multicolumn{2}{c}{Move vs Move}&\multicolumn{3}{c}{Average} \\ 
\cline{2-10}

~ & Acc6 & Acc7 & Acc6 & Acc7 & Acc6 & Acc7 & Acc6 & Acc7 & Runtime(ms)\\
\midrule
LPRNet\textsuperscript{a}\cite{zherzdev2018lprnet} & 77.68 & 75.67 & 47.73 & 41.26 & 61.43 & 57.06&63.86&59.81&\underline{0.33} \\
Eulpr\cite{qin2020efficient} & 81.22 & 79.83 & 64.47 & 61.01 & 65.95 & 62.85&70.01&65.62&0.65 \\
\bottomrule
CRNN\cite{shao2023multi} & 72.94&71.37&51.20&45.37&65.14&62.55&64.54&61.57&0.6 \\
LPRNet\textsuperscript{b}\cite{zherzdev2018lprnet} & 74.12&71.85&48.79&44.51&62.1&59.38&62.89&60.03&\textbf{0.26} \\
DAN\cite{xiao2021robust} & 78.17&76.34&57.47&54.39&\underline{73.18}&\underline{71.62}&71.35&69.40&1.7 \\
MFLPR-Net\cite{wang2022lsv}&\underline{80.29}&\underline{78.57}&\textbf{71.21}&\textbf{69.23}&\textbf{75.5}&\textbf{74.31}&\textbf{75.99}&\textbf{74.49}&1.8\\
LPTR-AFLNet&\textbf{82.67}&\textbf{80.22}&\underline{70.57}&\underline{67.16}&69.98&67.59&\underline{74.10}&\underline{71.50}&0.45 \\
\bottomrule
\end{tabular*}
\begin{tablenotes}   
    \footnotesize               
    \item[1]{ The hardware environment for reference \cite{wang2022lsv} was an Intel Core 3.4 GHz CPU with 12 GB RAM and four NVIDIA 1080Ti GPUs. Our experimental environment utilized an Intel Xeon 2.1 GHz CPU with 80 GB RAM and TITAN X GPUs. In the table, 'LPRNet\textsuperscript{a}' denotes the results from our trained model, while 'LPRNet\textsuperscript{b}' refers to the results from the model trained in the LSV dataset reference \cite{wang2022lsv}.}
\end{tablenotes}            
    \end{threeparttable}
\end{table}

Due to the presence of a large number of blurry license plate images in the LSV dataset, particularly in the "move vs static" and "move vs move" subgroups, existing license plate recognition algorithms generally demonstrate low accuracy on this dataset. Nevertheless, as shown in Table\ref{tbl3}, our proposed method maintains strong competitiveness in both recognition speed and accuracy. Compared to MFLPR-Net, our approach performs better on the "static vs move" subgroup, primarily benefiting from the relatively clearer license plate images within this subset. For such images, the model we introduce can quickly extract license plate features and, after warping the plates to a frontal view, passes the results to subsequent recognition networks.

However, on other subsets, our method's accuracy is 3 to 7 percentage points lower than that of MFLPR-Net. This is mainly due to the higher level of blur in the images of these subsets, which diminishes the feature extraction capability of lightweight recognition models compared to heavier models, leading to decreased performance. Additionally, although "static vs move" and "move vs static" are both single-sided motion scenarios, the "move vs static" condition involves camera movement, which introduces shake and results in lower image quality compared to "static vs move," thus reducing recognition accuracy. Interestingly, all methods achieve higher recognition rates in the "move vs move" scenario than in "move vs static." This may be because the motion states in both moving scenarios are more similar, resulting in data distributions that are closer, and consequently, better recognition performance.
\begin{table}
\caption{Inference Speed Comparison Across Different Hardware Platforms}\label{tbl4}
\begin{tabular*}{\tblwidth}{@{}CCCCC@{}}
\toprule
Equipment type&RAM&FPS&Hard disk type&Hard disk model \\ 
\bottomrule
TITAN X GPU+ Intel Xeno E5-2620 CPU&80&107&mechanical drive&ST4000NM0033-9ZM170\\
Intel Xeno E5-2620 CPU&80&79&mechanical drive&ST4000NM0033-9ZM170\\
Intel Core i7-7700K CPU&24&50&mechanical drive&ST1000DM003-1CH162\\
Intel Core i7-4770K CPU&32&33&solid state drive&KINGSTON SA400S37480G\\
\bottomrule
\end{tabular*}
\end{table}

To validate the feasibility of our proposed lightweight model in real-world deployment, we conducted inference speed tests across various hardware environments, with the results presented in Table\ref{tbl4}. As shown in Table\ref{tbl4}, even on standard CPU devices, our model can meet real-time requirements, further demonstrating its promising potential for practical applications.
\subsubsection{Ablation Evaluation}\label{subsubsec:Ablation-Evaluation}
To evaluate the individual contributions of each module within the proposed algorithm, we employed the aforementioned rectangular bounding box random perturbation strategy to simulate license plate localization results. Ablation experiments were then conducted on the Base subset of the CCPD dataset and the Move and Static subsets of the LSV dataset. Specifically, we statistically analyzed the recognition performance of the following model combinations: LPRNet, PTR+LPRNet, PTR+LPRNet+LP-CA, PTR+LPRNet+Focal CTC, and PTR+LPRNet+LP-CA+Focal CTC. To comprehensively assess the experimental effectiveness, in addition to the aforementioned Accuracy\_7c and Accuracy\_6c, we introduced a Character-level Precision (CP) metric. The latter represents the ratio of correctly recognized license plate characters to the total number of license plate characters, serving as a character-based measure of model recognition performance.
\begin{table}
\caption{Ablation study on CCPD dataset (\%)}\label{tbl5}
\begin{tabular*}{\tblwidth}{@{}CCCCC@{}}
\toprule
Ablation Experimen&Accuracy\_7c&Accuracy\_6c&CP\\ 
\bottomrule
LPRNet&	96.58	&97.35	&99.73\\
PTR+LPRNet&	99.00	&99.23&	99.88\\
PTR+LPRNet+LP-CA	&99.39	&99.68&	99.92\\
PTR+LPRNet+Focal CTC Loss	&99.35	&99.44	&99.92\\
PTR+LPRNet+LP-CA +Focal CTC Loss&	99.46	&99.56&	99.92\\
\bottomrule
\end{tabular*}
\end{table}
\begin{table}
\caption{Ablation study on LSV dataset (\%)}\label{tbl6}
\begin{tabular*}{\tblwidth}{@{}CCCCC@{}}
\toprule
Ablation Experimen&Accuracy\_7c&Accuracy\_6c&CP\\ 
\bottomrule
LPRNet	&41.26&	47.73	&87.23\\
PTR+LPRNet	&58.19	&61.57&	90.36\\
PTR+LPRNet+LP-CA&	62.51&	65.46&	91.76\\
PTR+LPRNet+Focal CTC Loss&	59.75&	63.70&	91.42\\
PTR+LPRNet+LP-CA+Focal CTC Loss&	67.16&	70.57	&93.92\\
\bottomrule
\end{tabular*}
\end{table}

As shown in Table\ref{tbl5} and Table\ref{tbl6}, the experimental results demonstrate that the proposed PTR, Focal CTC Loss, and LP-CA modules all contribute positively to improving the recognition rate, with PTR exhibiting the most significant impact. Building upon PTR, the effects of LP-CA and Focal CTC Loss are comparable, potentially due to their shared mechanism of enhancing the model's attention towards different targets through weighted approaches. Notably, the combination of all three modules on the CCPD dataset results in a substantial improvement of 2.88\% and 2.21\% in Accuracy\_7c and Accuracy\_6c, respectively. Since Accuracy\_7c accounts for the accuracy of all characters in the license plates, while Accuracy\_6c considers only the six characters excluding the Chinese character, Focal CTC Loss effectively mitigates the error rate of Chinese character recognition, which is often exacerbated by data imbalance issues. This leads to a slightly greater improvement in Accuracy\_7c compared to Accuracy\_6c. The character-level precision (CP) focuses on the recognition of every character. Due to the large number of license plates and the substantial character base, the improvement in CP (0.19\%) is relatively modest compared to Accuracy\_7c and Accuracy\_6c. Furthermore, considering the sheer volume of characters, even subtle differences in Accuracy\_7c between the PTR+LPRNet+LP-CA and PTR+LPRNet+Focal CTC combinations are unlikely to be reflected in the two-decimal-place precision of the CP metric. The experimental findings on the LSV dataset align with those on the CCPD dataset. Specifically, Accuracy\_7c and Accuracy\_6c are improved by 25.90\% and 22.84\%, respectively, while CP increases by 6.69\%.
\begin{table}[htbp]
\centering
\caption{License Plate Recognition Error Statistics on CCPD (\%)}\label{tbl7}
\begin{tabularx}{\textwidth}{@{}>{\centering\arraybackslash}X *{6}{>{\centering\arraybackslash}c}@{}}
\toprule
Model & Special Category & Chinese Chars & Mis-Increase & Missing Chars & Confusion & Total \\ 
\midrule
LPRNet & 2 & 2661 & 6100 & 4378 & 2519 & 15660 \\
PTR+LPRNet+LP-CA+Focal CTC & \multirow{2}*{2} &\multirow{2}*{648} & \multirow{2}*{2084} & \multirow{2}*{960} & \multirow{2}*{340} & \multirow{2}*{4034} \\
\bottomrule
\end{tabularx}
\vspace{2mm}
\end{table}
\begin{figure}
  \centering
    \includegraphics[width=0.8\linewidth]{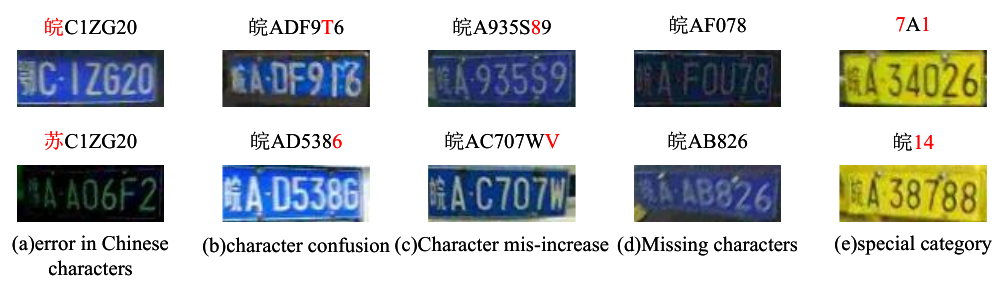}
    \caption{Sample images of license plates with recognition errors}\label{fig29}
\end{figure}

\begin{figure}
  \centering
    \includegraphics[width=0.8\linewidth]{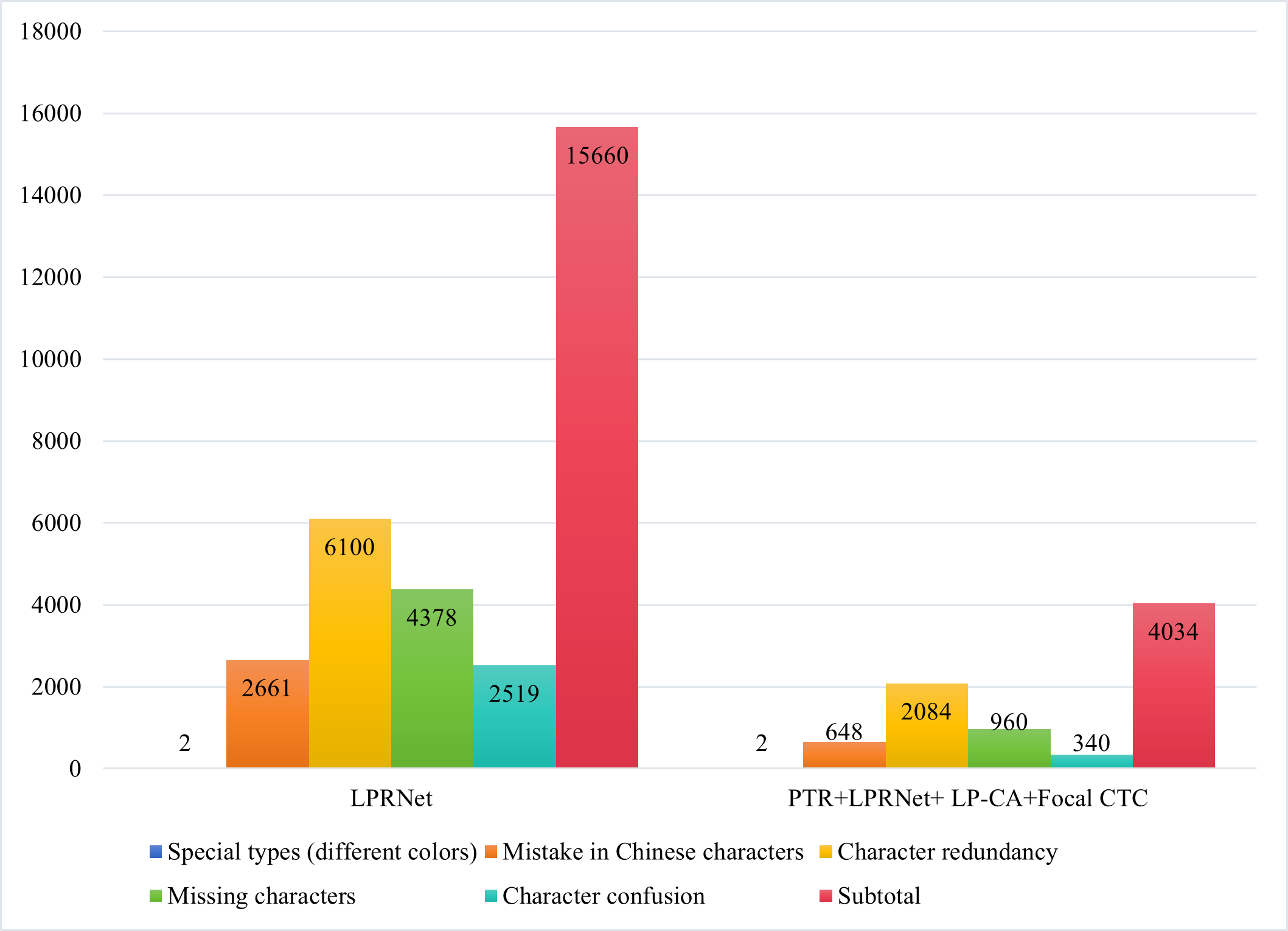}
    \caption{Bar Graph Visualization of error statistics for license plate types on the CCPD dataset}\label{fig30}
\end{figure}

A statistical analysis was performed to evaluate the license plate image recognition errors before and after the proposed model improvements. The results are summarized in Table\ref{tbl7} Misidentified license plates were categorized into five types: special plate types, Chinese character errors, superfluous characters, missing characters, and character confusion. Representative examples of each error category are illustrated in Figure\ref{fig29}. To facilitate a more intuitive comparison, the data presented in Table\ref{tbl7} were visualized as a histogram in Figure\ref{fig30}. As demonstrated in Figure\ref{fig30}, the implemented model improvements demonstrably alleviated the identified error categories, resulting in a significant reduction in the total number of misidentified license plate images. However, the error rate associated with special plate types did not exhibit a corresponding significant improvement. This lack of improvement is primarily attributed to the absence of license plates with non-standard colors in the training dataset.
\subsection{Qualitative Error Analysis and Visualizations}\label{subsec:Qualitative-Error-Analysis-and-Visualizations}
\begin{figure}
  \centering
    \includegraphics[width=0.8\linewidth]{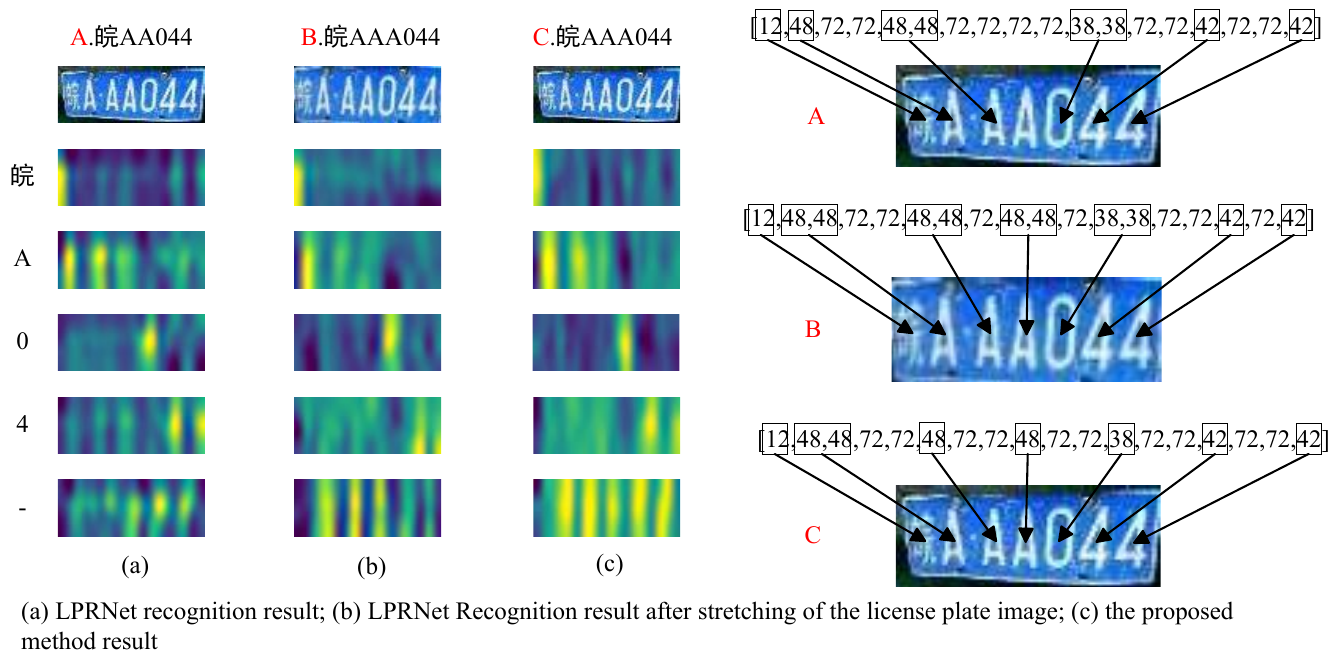}
    \caption{Qualitative Error Analysis and Visualizations of Character Omission Errors}\label{fig31}
\end{figure}
To further elucidate the underlying reasons for the effectiveness of the proposed improved algorithm, we performed a visualization and analysis of error cases based on activation maps of intermediate network layers and the corresponding character IDs from CTC decoding outputs. Figure\ref{fig31} illustrates a scenario where confounding factors weaken character features, leading to the erroneous omission of the character 'A' in the recognition result. On the left side, (a), (b), and (c) depict the 73 × 4 × 18 feature maps output by the network's intermediate layers under three different conditions. On the right side, A, B, and C indicate the character IDs obtained from CTC decoding in each respective case. In Figure\ref{fig31}(a), the output of the original LPRNet is shown, where the letter 'A' is mistakenly omitted—a problem that is analyzed in detail in Section \ref{subsubsec:LPRNet-Optimization-with-LP-CA}. We hypothesize that this issue arises from mutual interference among character features.

To validate our hypothesis, we kept the model parameters unchanged and horizontally stretched the input images to increase the spacing between characters. We then tested again using LPRNet, and the results showed correct recognition (as shown in Figure\ref{fig31}(b)). This indicates that increasing the inter-character spacing reduces interference between character features, allowing the model to better recognize all characters.

The proposed Lightweight Per-Channel Attention (LP-CA) module allows each high-level feature channel in the network to adaptively interact with features from its left and right neighbors based on its type, via a learned 1×3 convolutional kernel, thereby improving the accuracy of subsequent recognition. As shown in Figure\ref{fig31}(c), LP-CA significantly enhances the semantic information of various inter-character spacing locations within the spacing dictionary page compared to (a). Furthermore, the features of the third 'A' character in the corresponding dictionary page are also slightly strengthened. Simultaneously, the activation values at corresponding positions within the spacing dictionary page are attenuated. Consequently, even if the activation intensity of this particular 'A' character is weaker than other 'A' characters, it can still be correctly recognized.

\begin{figure}
  \centering
    \includegraphics[width=0.8\linewidth]{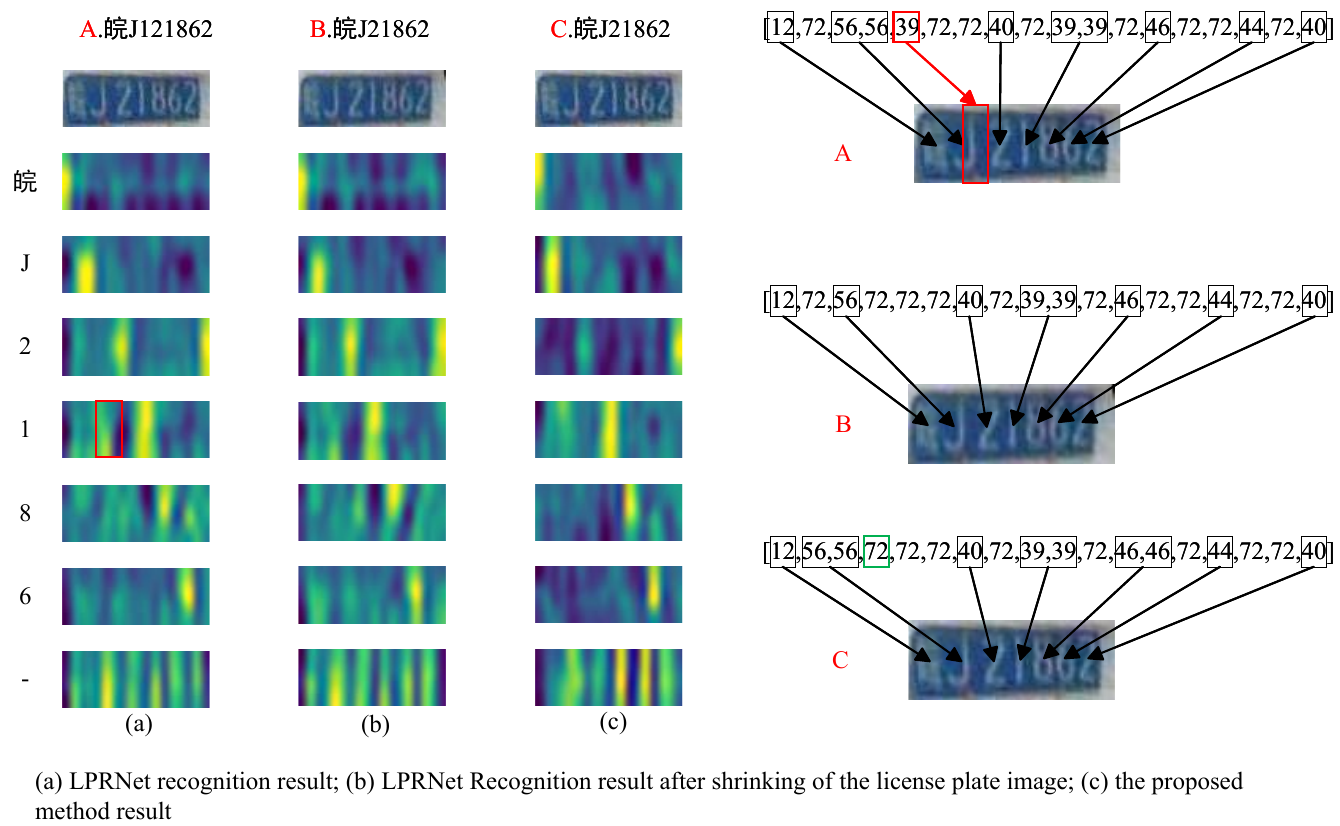}
    \caption{Qualitative Error Analysis and Visualizations of Character Insertion Errors}\label{fig32}
\end{figure}

 Figure\ref{fig32} illustrates instances of character over-insertion errors. As shown in (a), the character ‘J’ bears a glyph shape that resembles ‘1’ on its right side; consequently, in the response at the corresponding ‘1’ dictionary page location (indicated by the red box in (a)), the response underneath is slightly amplified. Meanwhile, the response of the ‘-’ entry at that position in the dictionary page is somewhat weakened, leading to a misrecognition of ‘1’, with its predicted ID being 39, as marked by the red box in the upper right corner of Figure\ref{fig32}. When the image is compressed horizontally, as shown in (b), the spatial area shrinks, causing the feature region of ‘J’ to diminish. This results in a reduced activation response for ‘1’ in the corresponding dictionary page within LPRNet, while the response for ‘-’ at that position is enhanced. As a consequence, the ‘-’ is correctly identified as a separator, as indicated by the green box in the lower right C subfigure.
 
After the implementation of the proposed per-channel lightweight attention module, LP-CA, as shown in Figure\ref{fig32}(c), the activation values of all dictionary pages except the gap dictionary page have been significantly enhanced. Although the responses at the four positions on the right side of the gap dictionary page are also improved and tend to be more concentrated, the responses at the first two positions remain relatively weak and somewhat dispersed. Notably, due to the elevated activation values across other dictionary pages, the overall recognition results remain accurate despite the less-than-ideal responses on the gap dictionary page. This robustness is attributed to the fact that the parameters of the attention module are optimized and learned over the entire dataset, endowing the model with strong generalization capability.
\begin{figure}
  \centering
    \includegraphics[width=0.4\linewidth]{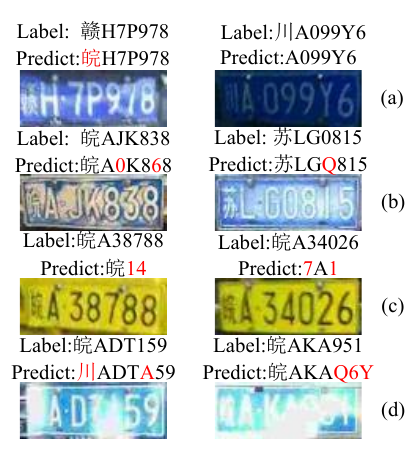}
    \caption{Examples of incorrect license plate recognition in specific scenarios}\label{fig33}
\end{figure}

Undeniably, the proposed algorithm still encounters errors when recognizing license plates under specific circumstances. These errors primarily fall into four categories:

First, as illustrated in Figure\ref{fig33}(a), the model continues to exhibit errors in the recognition of Chinese characters. For instance, confusion frequently arises when identifying Chinese characters from provinces other than Anhui (“WAN”). This issue is primarily attributed to the imbalanced distribution of Chinese character samples within the CCPD dataset. Specifically, license plates from Anhui Province are over represented, while those from other provinces are relatively scarce. Although Focal Loss mitigates this problem to a certain extent, it does not provide a complete solution.

Second, as shown in Figure\ref{fig33}(b), ambiguity and interference between similar characters on license plates can also trigger recognition errors. This is mainly due to factors like paint chipping or glare, resulting in the background color becoming whitish. Consequently, the white characters merge with the whitened background, causing false activations during feature detection. For example, the letter ‘J’ and the digit ‘0’ share considerable visual similarity, particularly in their lower right portions. Combined with the whitish background on the left, this can lead the model to incorrectly identify the ‘J’ in the first license plate in Figure (b) as a ‘0’. Similarly, the misidentification of ‘0’ as ‘Q’ in the second license plate in Figure (b) arises from a comparable situation. The whitening in the lower right part of the ‘0’ creates a prominent white blotch, prompting the erroneous classification.

Third, as evidenced in Figure\ref{fig33}(c), the model struggles to accurately recognize yellow single-line license plates. The primary reason for this difficulty is the predominance of blue license plates within the CCPD dataset. The limited availability of license plates with other colors renders the model overly sensitive to color and hinders its ability to generalize to license plates of different hues effectively.

Finally, the fourth scenario, depicted in Figure\ref{fig33}(d), stems from the poor quality of license plate images where the characters are blurred and even challenging to discern by the human eye. The indistinct character features in these low-quality images significantly compromise the model's recognition performance, resulting in incorrect identification outcomes.
\section{Conclusion}\label{sec:Conclusion}
This paper introduces LPTR-AFLNet, a lightweight, unified network for both correcting and recognizing single/double-line Chinese license plates. The network innovatively leverages a Perspective Transformation Rectification (PTR) module, effectively overcoming the limitations of traditional Spatial Transformer Networks (STN) in license plate perspective rectification. Furthermore, addressing the shortcomings of LPRNet, the incorporation of a custom, lightweight per-channel attention module (LP-CA) and Focal CTC loss significantly improves recognition accuracy. The extended PTR module not only automates the rectification and concatenation of the upper and lower character regions of double-line license plates but also benefits from joint optimization with the improved LPRNet, enabling unified correction and recognition of both single and double-line Chinese license plates. Experimental results clearly demonstrate that LPTR-AFLNet excels in handling single/double-line license plate images with perspective distortion, maintaining high recognition accuracy even in various complex scenarios while preserving real-time performance.

Despite the achievements of the current algorithm, there is still room for improvement. Future research could focus on several key areas: Firstly, a strong emphasis should be placed on effective data augmentation to construct a more comprehensive and larger dataset, thereby enhancing the model's generalization capability. Secondly, exploring the integration of an image quality assessment module as a pre-processing step in the recognition pipeline, filtering out low-quality license plate images, has the potential to improve overall recognition accuracy. Finally, further optimization of the PTR module's network architecture is warranted to achieve superior rectification performance while maintaining speed advantages.

\clearpage 




\bibliographystyle{cas-model2-names}

\bibliography{cas-refs}



\end{document}